\documentclass[final]{cvpr}
\usepackage{nopageno}

\usepackage{times}
\usepackage{epsfig}
\usepackage{graphicx}
\usepackage{amsmath}
\usepackage{amssymb}
\usepackage{booktabs}

\usepackage{xcolor}
\usepackage{tablefootnote}
\usepackage{footnote}
\usepackage[switch]{lineno} %
\usepackage{graphicx}
\usepackage{subcaption}
\usepackage{multirow}
\usepackage[export]{adjustbox}
\usepackage[toc,page]{appendix}
\usepackage{graphicx,floatrow}

\DeclareMathOperator*{\argmax}{argmax} 
\DeclareMathOperator*{\argmin}{argmin} 

\usepackage[pagebackref=true,colorlinks,bookmarks=false,breaklinks=true]{hyperref}

\def\cvprPaperID{7308} 
\def\confYear{CVPR 2021}
\begin{document}

\title{Over-the-Air Adversarial Flickering Attacks against Video Recognition Networks}

\author{Roi Pony\textsuperscript{\rm 1}\thanks{Equal contribution} ,
        Itay Naeh\textsuperscript{\rm 2}\footnotemark[1] , 
        Shie Mannor\textsuperscript{\rm 1,3}\\\\
        \small
        \textsuperscript{\rm 1}Department of Electrical Engineering,
Technion Institute of Technology, Haifa, Israel \\
    \small\textsuperscript{\rm 2} Rafael - Advanced Defense Systems Ltd., Israel \\
        \small\textsuperscript{\rm 3} Nvidia Research \\
    \tt\small roipony@gmail.com, itay@naeh.us, shie@technion.ac.il
    
}

\maketitle

\begin{abstract}
Deep neural networks for video classification, just like image classification networks, may be subjected to adversarial manipulation. The main difference between image classifiers and video classifiers is that the latter usually use temporal information contained within the video. In this work we present a manipulation scheme for fooling video classifiers by introducing a flickering temporal perturbation that in some cases may be unnoticeable by human observers and is implementable in the real world. After demonstrating the manipulation of action classification of single videos, we generalize the procedure to make universal adversarial perturbation, achieving high fooling ratio. In addition, we generalize the universal perturbation and produce a temporal-invariant perturbation, which can be applied to the video without synchronizing the perturbation to the input. The attack was implemented on several target models and the transferability of the attack was demonstrated. These properties allow us to bridge the gap between simulated environment and real-world application, as will be demonstrated in this paper for the first time for an over-the-air flickering attack.
\end{abstract}

\begin{figure}

\begin{minipage}[b]{\columnwidth}
\centering
\begin{subfigure}[b]{\linewidth}
 \centering
\includegraphics[width=\linewidth,trim=0.15cm 0.5cm 0cm 0.5cm,clip]{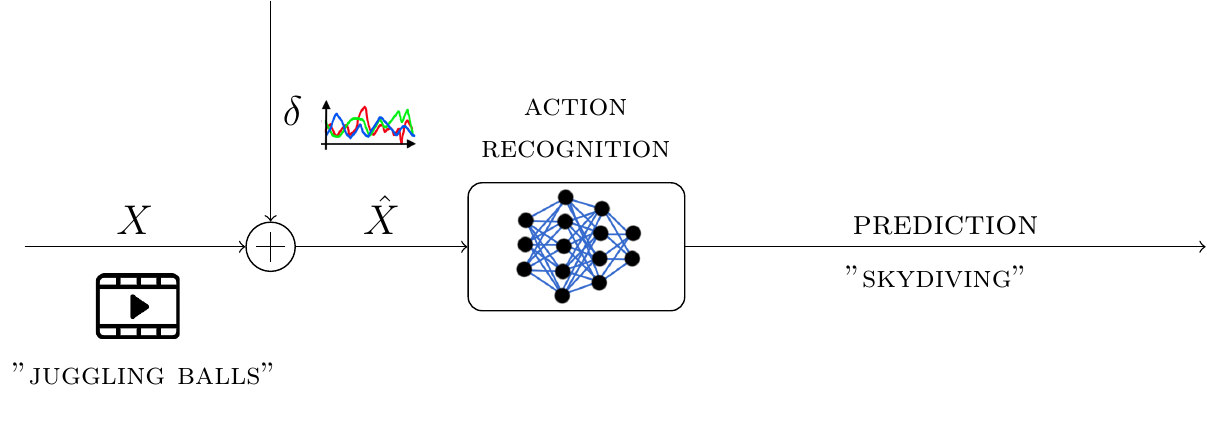}
\caption{Diagram of a Flickering Adversarial Attack in a simulated environment (digital).}
\label{fig:overview_digital}
\vspace{5mm}
\end{subfigure}
\begin{subfigure}[b]{\linewidth}
 \centering
\includegraphics[width=\linewidth,trim=0.15cm 0.5cm 0cm 0.5cm,clip]{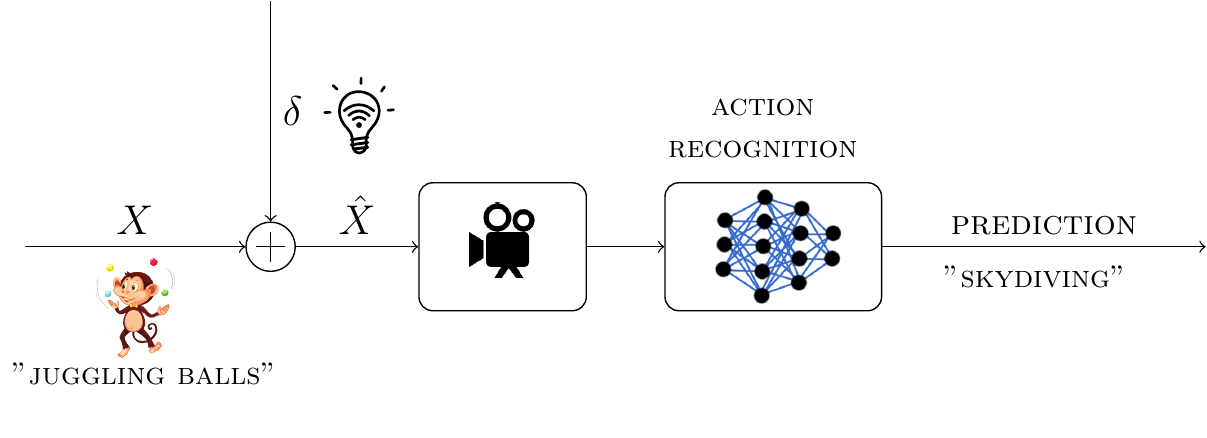}
\caption{Diagram of an Over-the-Air Flickering Adversarial Attack in the real-world (physical).}
\label{fig:overview_ota}
\end{subfigure}
\end{minipage}
\caption{Top figure shows the attack diagram in the digital domain performed by adding a uniform RGB perturbation to the attacked video. Bottom figure shows the modeling of the digitally-developed attack into the real-world by transmitting the perturbation in the scene using a smart RGB led bulb.}
\label{fig:overview_digital_and_ota}
\end{figure}

\section{Introduction}
In recent years, Deep Neural Networks (DNNs)  have shown phenomenal performance in a wide range of tasks, such as image classification \cite{NIPS2012_4824}, object detection \cite{NIPS2015_5638}, semantic segmentation \cite{DBLP:journals/pami/ShelhamerLD17} etc.
Despite their success, DNNs have been found vulnerable to adversarial attacks.
Many works \cite{Szegedy14,Goodfellow15,papernot2016limitations} have shown that a small (sometimes imperceptible) perturbation added to an image, can make a given DNNs prediction false.
These findings have raised many concerns, particularly for critical systems such as face recognition systems \cite{NIPS2014_5416}, surveillance cameras \cite{Sultani_2018_CVPR},  autonomous vehicles, and medical applications \cite{LITJENS201760}. In recent years most of the attention was given to the study of adversarial patterns in images and less in video action recognition. Only in the past two years works on adversarial video attacks were published \cite{WeiS19,Inkawhich_18,Wei2019BlackboxAA,Jiang2019BlackboxAA}, even though DNNs have been applied to video-based tasks for several years, in particular video action recognition \cite{Zisserman17,Wang2017NonlocalNN,Feichtenhofer18}. In video action recognition networks temporal information is of the essence in categorizing actions, in addition to per-frame image classification. In some of the proposed attacks the emphasis was, beyond adversarial categorization, the sparsity of the perturbation. In our work, we consider adversarial attacks against video action recognition under a white-box setting, with an emphasis on the imperceptible nature of the perturbation in the spatio-temporal domain to the human observer and implementability of the generalized adversarial perturbation in the real-world. 
We introduce flickering perturbations by applying a uniform RGB perturbation to each frame, thus constructing a {\it temporal} adversarial pattern. 
Unlike previous works, in our case sparsity of the pattern is undesirable, because it helps the adversarial perturbation to be detectable by human observers for its unnatural pattern, and to image based adversarial perturbation detectors for the exact same reason.
The adversarial perturbation presented in this work does not contain any spatial information on a single frame other than a constant offset. This type of perturbation often occurs in natural videos by changing lighting conditions, scene changes, etc.
In this paper, we aim to attack the video action recognition task \cite{Will17_kinetics}.
For the targeted model we focus on the I3D \cite{Zisserman17} model (Specifically we attack the RGB stream of the model, rather than on the easier to influence optical flow stream) based on InceptionV1 \cite{Szegedy15_Inceptionv1} and we expand our experiments to additional models from \cite{tran2018closer}. The attacked models trained on the Kinetics-400 Human Action Video Dataset \cite{Will17_kinetics}. 

In order to make the adversarial perturbation unnoticeable by human observers, we reduce the thickness and temporal roughness of the adversarial perturbation, which will be defined later in this paper. In order to do so we apply two regularization terms during the optimization process, each corresponds to a different effect of the perceptibly of the adversarial pattern.
In addition, we introduce a modified adversarial-loss function that allows better integration of these regularization terms with the adversarial loss.

We will first focus on the I3D \cite{Zisserman17} network and introduce a flickering attack on a single video and present the trade-off between the different regularization terms. We construct universal perturbations that generalize over classes and achieve $93\%$ fooling ratio. Another significant feature of our proposed method is time invariant perturbations that can be applied to the classifier without synchronization. This makes the perturbation relevant for real world scenarios, since frame synchronization is rarely possible. We show the effectiveness of the flickering attack on other models \cite{tran2018closer} and the inter-model transferability, and finally demonstrate the over-the-air flickering attack in a real world scenario for the first time. A diagram of the digital attack and the over-the-air attack pipelines are shown in Figure \ref{fig:overview_digital_and_ota}.\\
The main contributions of this work are:
\begin{itemize}
    \item A methodology for developing flickering adversarial attacks against video action recognition networks that incorporates a new type of regularization for affecting the visibility of the adversarial pattern.
    \item A universal time-invariant adversarial perturbation that does not require frame synchronization.
    \item Adversarial attacks that are transferable between different networks. 
    \item Adversarial attacks that are implementable using temporal perturbations.
\end{itemize}  

The paper is organized as follows: We briefly review related work and present the flickering adversarial attack. Then we show experimental results and the generalization of the attack. Finally, we present real world examples of the flickering adversarial attacks, followed by conclusions and future work.
We encourage the readers to view the attack videos\footnote{\label{youtube_videos} \url{https://bit.ly/Flickering_Attack_videos}}, over-the-air scene-based attack videos\footnote{\url{https://bit.ly/Over_the_Air_scene_based_videos}\label{ota_scene_youtube_videos}}, and over-the-air universal attack videos\footnote{\label{ota_youtube_videos}\url{https://bit.ly/Over_the_Air_videos}}.
Our code can be found in the following repository\footnote{\raggedright \url{https://bit.ly/Flickering_Attack_Code}}.


\section{Related Work}\label{Related Work}
\subsection{Video Action Recognition}
With deep Convolutional Neural Networks (CNNs) achieving state-of-the-art performance on image recognition tasks, many works propose to adapt this achievement to video-based computer vision tasks. 
The most straightforward approach for achieving this is to add temporally-recurrent layers such as LSTM \cite{SakSB14_LSTM} models to  traditional 2D-CNNs. This way, long-term temporal dependencies can be assigned to spatial features \cite{Wang18Human_Action_Recognition,SimonyanZ14_two_stream_conv}.
Another approach implemented in C3D \cite{JiXYY13_C3D,TranBFTP14,VarolLS16_Long_term_Temporal} extends the 2D CNNs (image-based) to 3D CNNs (video-based) kernels and learns hierarchical spatio-temporal representations directly from raw videos.
Despite the simplicity of this approach, it is very difficult to train such networks due to their huge parameter space.
To address this, \cite{Zisserman17} proposes the Inflated 3D CNN (I3D) with inflated 2D pre-trained filters \cite{RussakovskyDSKSMHKKBBF14_imageNet}.
In addition to the RGB pipeline,  optical flow is also useful for temporal information encoding, and indeed several architectures greatly improved their performance by incorporating an optical-flow stream \cite{Zisserman17}. 
\cite{tran2018closer} demonstrated the advantages of 3D CNNs over 2D CNNs within the framework of residual learning, proposing factorization of the 3D convolutional filters into separate spatial and temporal components.
\subsection{Adversarial Attack on Video Models}
The research of the vulnerability of video-based classifiers to adversarial attacks emerged only in the past years. The following attacks were performed under the white-box attack settings:
\cite{WeiS19} were the first to investigate a white-box attack on video action recognition. They proposed an $L_{2,1}$ norm based optimization algorithm to compute sparse adversarial perturbations, focusing on networks with a CNN+RNN architecture in order to investigate the propagation properties of perturbations.
\cite{LiNPSKRS19} generated an offline universal perturbation using a GAN-based model that they applied to the learned model on unseen input for real-time video recognition models.
\cite{Roberto_18}
proposed a nonlinear adversarial perturbation by using another neural network model (besides the attacked model), which was optimized to transform the input into adversarial pattern under the $L_{1}$ norm.
\cite{Inkawhich_18} proposed both white and black box untargeted attacks on two-stream model (optical-flow and RGB), based on the original and the iterative version of FGSM \cite{Goodfellow15,KurakinGB17}, and used FlowNet2 \cite{IlgMSKDB16_flowNet2} to estimate optical flow in order to provide gradients estimation.
Several black-box attacks were proposed \cite{Jiang2019BlackboxAA,Wei2019BlackboxAA}. Our attack follows the white-box setting therefore those attacks are beyond the scope of this paper.
\section{Flickering Adversarial Attack}\label{Flickering Adversarial attack}
The flickering adversarial attack consists of a uniform offset added to the entire frame that changes each frame. This novel approach is desirable for several reasons. First, it contains no spatial pattern within individual frames but an RGB offset. Second, this type of perturbation can easily be mistaken in some cases as changing lighting conditions of the scene or typical sensor behaviour. Third, it is implementable in the real-world using a simple LED light source.
\subsection{Preliminaries}
 Video action recognition is a function $F_{\theta}(X)=y$  that accepts an input $X =[x_{1}, x_{2},..,x_{T}] \in \mathbb{R}^{T\times H \times W \times C}$ from $T$ consecutive frames with $H$ rows, $W$ columns and $C$ color channels, and produces an output $y \in \mathbb{R}^{K}$ which can be treated as probability distribution over the output domain, where $K$ is the number of classes. The model $F$ implicitly depends on some  parameters $\theta$ that are fixed during the attack. The classifier assigns the label $A_{\theta}(X)=\argmax_{i}y_{i}$ to the input $X$.
We denote adversarial video by $\hat{X} = X+\delta$ where the video perturbation $\delta =[\delta_{1}, \delta_{2},..,\delta_{T}]\in \mathbb{R}^{T\times H \times W \times C}$, and each individual adversarial frame by  $\hat{x}_i = x_i +\delta_{i}$. $\hat{X}$ is adversarial when $A_{\theta}(\hat{X}) \neq A_{\theta}(X)$ (untargeted attack) or $A_{\theta}(\hat{X}) = k \neq A_{\theta}(X)$ for a specific predetermined incorrect class $k\in \left[K\right]$ (targeted attack), while keeping the distance between $\hat{X}$ and ${X}$ as small as possible under the selected metric (e.g., $L_{2}$ norm). 
\subsection{Methodology}
In our attack $\delta_{i}$ is designed to be spatial-constant on the three color channels of the frame, meaning that for each pixel in image $x_{i}$, an offset is added with the same value (RGB). Thus, the $i^{th}$ perturbation $\delta_{i}$, which corresponds to the $i^{th}$ frame $x_{i}$ of the video, can be represented by three scalars, hence $\delta=[\delta_{1}, \delta_{2},..,\delta_{T}]\in \mathbb{R}^{T\times 1 \times 1 \times 3}$, having in total $3T$ parameter to optimize.
 To generate an adversarial perturbation we usually use the following objective function
\begin{alignat}{2}
\label{eq:opt_eq_gen}
&\operatornamewithlimits{\argmin_{\delta}} \lambda\sum_{j}{ \beta_{j}D_{j}(\delta)} + \frac{1}{N}\sum_{n=1}^{N}{\ell(F_{\theta}(X_{n}+\delta),t_{n})}\\
\label{eq:constrain_gen}& s.t \ \hat{x}_i \in [V_{min}, V_{max}]^{H \times W \times C},
\end{alignat}
where $N$ is the total number of training videos, $X_{n}$ is the $n^{th}$ video, $F_{\theta}(X_{n}+\delta)$ is the classifier output (probability distribution or logits), and $t_{n}$ is the original label (in the case of untargeted attack).
The first term in Equation (\ref{eq:opt_eq_gen}) is regularization term, while the second is adversarial classification loss, as will be discussed later in this paper.
The parameter $\lambda$ weighs the relative importance of being adversarial and also the regularization terms.
The set of functions $D_{j}(\cdot)$ controls the regularization terms that allows us to achieve better imperceptibility for the human observer. The parameter $\beta_{j}$ weighs the relative importance of each regularization term.
The constraint in Equation (\ref{eq:constrain_gen}) guarantees that after applying the adversarial perturbation, the perturbed video will be clipped between the valid values: $V_{min},V_{max}$, that represents the minimum and maximum allowed pixel intensity.
\subsection{Adversarial loss function}\label{Modified Adversarial loss function}
We use a loss mechanism similar to the loss presented by C\&W \cite{CarliniW16a}, with a minor modification.
For untargeted attack:
\begin{equation}\label{eq:adv_loss}
\ell(y,t) =
\max \left(0,  \min\left(\frac{1}{m} \ell_{m}(y,t)^2,
\ell_{m}(y,t)\right) \right)
\end{equation}
\begin{equation}\label{eq:sub_adv_loss}
\ell_{m}(y,t) = y_{t}- \max_{i\neq t}(y_{i}) +m,
\end{equation}
where $m>0$ is the desired margin of the original class probability below the adversarial class probability.
A more detailed explanation of the motivation in defining the above loss function is found in the supplementary material.

\subsection{Regularization terms}\label{Regularization terms}
We quantify the distortion introduced by the perturbation $\delta$ with $D(\delta)$ in the spatio-temporal domain. This metric will be constrained in order for the perturbation $\delta$ to be imperceptible to the human observer while remaining adversarial. Unlike previously published works on adversarial patches in images, in the video domain imperceptible may reference thin patches in gray-level space or slow changing patches in temporal frame space.
In contrast to previous related works \cite{WeiS19,Wei2019BlackboxAA}, in our case temporal sparsity is not of the essence but the unnoticability to the human observer. In order to achieve the most imperceptible perturbation we introduce two regularization terms, each controlling different aspects of human perception.

In order to simplify the definition of our regularization terms and metrics, we define the following notations for  $X =[x_{1}, x_{2},..,x_{T}] \in \mathbb{R}^{T\times H \times W \times C}$  (video or perturbation).\\
\textbf{\textit{Tensor p-norm:}} 
\begin{equation}\label{eq:p_norm}
\left\|X\right\|_{p} =\left(\sum_{i_{1}=1}^{T} \cdots \sum_{i_{4}=1}^{C}{|x_{i_{1} \ldots i_{4}}|^{p}}\right)^{1/p} ,
\end{equation}
\textit{where $i_{1},i_{2},.., i_{4}$ refer to dimensions.}\\
\textit{\textbf{Roll operator}: $Roll(X, \tau)$ produce the time shifted tensor, whose elements are $\tau$-cyclic shifted along the first axis (time)}:
\begin{equation}\label{eq:roll}
Roll(X, \tau)=[x_{\left(\tau\;\mathrm{mod}\;T\right)+1}, ...,x_{\left(T-1+\tau\;\mathrm{mod}\;T\right)+1}].
\end{equation}
\textit{\textbf{$1^{st}$ and $2^{nd}$ order temporal derivatives}:
We approximate the $1^{st}$ and $2^{nd}$ order temporal derivatives by finite differences as follows.}
\begin{align}\label{eq:1st_der}
\frac{\partial X}{\partial t}&= Roll(X, 1)- Roll(X, 0),
\end{align}
\begin{align}\label{eq:2st_der}
\frac{\partial^2 X}{\partial t^2}&= Roll(X, -1)- 2Roll(X, 0)+Roll(X, 1).
\end{align}
\subsubsection{Thickness regularization}
This loss term forces the adversarial perturbation to be as small as possible in gray-level over the three color channels (per-frame), having no temporal constraint and can be related to the ``thickness" of the adversarial pattern.
\begin{equation*}\label{eq:thick}
D_{1}(\delta)=\frac{1}{3T}\left\|\delta\right\|_{2}^{2},
\end{equation*}
where $\left\|\cdot\right\|_{2}$ defined in Equation (\ref{eq:p_norm}) with $p=2$.

\subsubsection{Roughness regularization}\label{Roughness_regularization}
We introduce temporal loss functions that incorporate two different terms,
\begin{equation}\label{eq:temporal}
D_{2}(\delta)=\frac{1}{3T}\left\|\frac{\partial \delta}{\partial t}\right\|_{2}^{2}+\frac{1}{3T}\left\|\frac{\partial^2 \delta}{\partial t^2}\right\|_{2}^{2},
\end{equation}
where $\frac{\partial \delta}{\partial t}$ and $\frac{\partial^2 \delta}{\partial t^2}$ are defined in Equations (\ref{eq:1st_der},\ref{eq:2st_der}), respectively.

The norm of the first order temporal difference shown in the Equation (\ref{eq:temporal}) (first term) controls the difference between each two consecutive frame perturbations. This term penalizes temporal changes of the adversarial pattern. Within the context of human visual perception, this term is perceived as ``flickering", thus we wish to minimize it.  

The norm of the second order temporal difference shown in Equation (\ref{eq:temporal}) (second term) controls the trend of the adversarial perturbation. Visually, this term  penalizes fast trend changes, such as spikes, and may be considered as scintillation reducing term.

The weights of $D_{1}$ and $D_{2}$ will be noted by $\beta_{1}$ and $\beta_{2}$, respectively, throughout the rest of the paper and also in the YouTube videos.

\subsection{Metrics}\label{Metric}
Let us define several metrics in order to quantify the performance of our adversarial attacks.\\
\textbf{\textit{Fooling ratio}}: is defined as the percentage of adversarial videos that are successfully misclassified (higher is better).
\textbf{\textit{Mean Absolute Perturbation per-pixel}}:
\begin{equation}\label{metric:thick}
thickness_{gl}(\delta) =\frac{1}{3T}\left\|\delta\right\|_{1},
\end{equation}
where $\left\|\cdot\right\|_{1}$ defined in Equation (\ref{eq:p_norm}) with $p=1$.\\
\textbf{\textit{Mean Absolute Temporal-diff Perturbation per-pixel}}:
\begin{equation}\label{metric:rough}
roughness_{gl}(\delta) =\frac{1}{3T}\left\|\frac{\partial \delta}{\partial t}\right\|_{1}.
\end{equation}
The thickness and roughness values in this paper will be presented as percents from the full applicable values of the image span, e.g.,
\begin{equation*}
thickness(\delta) = \frac{thickness_{gl}(\delta)}{V_{max}-V_{min}}*100.
\end{equation*}

\begin{table}
\caption{Results over several types of attacks on different attacked models. Thickness and Roughness defined in Equations (\ref{metric:thick},\ref{metric:rough})}
\label{table:result}
\begin{center}
\resizebox{1.\columnwidth}{!}{
\begin{tabular}{|c|c|c|c|c|c}
\hline
Attack  &Attacked Model     &Fooling ratio[\%] & Thickness[\%] &Roughness[\%] \\
\midrule
Single Video &I3D            & 100       & 1.0$\pm$0.5          & 0.83$\pm$ 0.4\\
Single Video &R(2+1)D                 & 93.0       & 2.4$\pm$1.9          & 2.1$\pm$ 2.0\\
\midrule
Single Class &I3D                & 90.2$\pm$ 11.72           & 13.0$\pm$ 3.6          & 10.6$\pm$ 2.2 \\
\midrule
Universal &I3D                       & 93.0            & 15.5           & 15.7 \\
Universal &R(2+1)D                      & 79.0            & 18.1           & 21.0 \\
Universal &MC3                     & 77.1            & 18.3           & 24.5 \\
Universal &R3D                    & 90.3            & 17.8           & 25.5 \\
\midrule
Universal Time Invariance  &I3D    & 83.1       & 18.0           & 14.0  \\
\bottomrule
\end{tabular}
 }
\end{center}
\end{table}

\section{Experiments on I3D}
\subsection{Targeted Model}
Our attack follows the white-box setting, which assumes the complete knowledge of the targeted model, its parameter values and architecture. The I3D \cite{Zisserman17} model for video recognition is used as target model, focused on the RGB pipeline. The adversarial attacks described in this work can be a targeted or untargeted, and the theory and implementation can be easily adapted accordingly. The I3D model was selected for targeting because common video classification networks are based upon its architecture. Therefore, the insights derived from this work will be relevant for these networks.
In the I3D configuration $T=90, H=224,W=224,C=3$, and  $V_{min}=-1, V_{max}=1$ (trained on the kinetics Dataset).
Implementation details can be found in the supplementary material.
\subsection{Dataset}
We use Kinetics-400 \cite{Will17_kinetics} for our experiments. Kinetics is a standard benchmark for
action recognition in videos. It contains about 275K video of 400 different human action categories (220K videos in the training split, 18K in the validation split, and 35K in the test split).  
For the single video attack we have developed the attacks using the validation set. In the class generalization section we  trained on the training set and evaluated on the validation set. In the universal attack section we trained on the validation set and evaluated on the test set.
We pre-processed the dataset by excluding movies in which the network misclassified to begin with and over-fitted entries.
Each video contains 90-frame snippets.
\subsection{Single Video Attack}\label{sec:single_vid_attack}
In order to perform the flickering adversarial attacks on single videos, a separate optimization
has to be solved for each video, i.e., solving Equation (\ref{eq:opt_eq_gen}) for a single video ($N=1$) s.t. each video have its own tailor-made $\delta$.
In our experiment we have developed different $\delta$'s for hundreds of randomly picked samples from the kinetics validation set.
The \textit{Single Video} entry in Table \ref{table:result} shows the statistics of average and standard deviation of the fooling ratio, thickness and roughness of untargeted single-video attacks, reaching $100\%$ fooling ratio with low roughness and thickness values.
Video examples of the attack can be found here\textsuperscript{\ref{youtube_videos}}.
Detailed description of the convergence process regarding this attack can be found in the supplementary material.
\begin{figure}
\centering
  {\includegraphics[width =\linewidth,trim=6.6cm 2.2cm 6cm 2.3cm,clip]{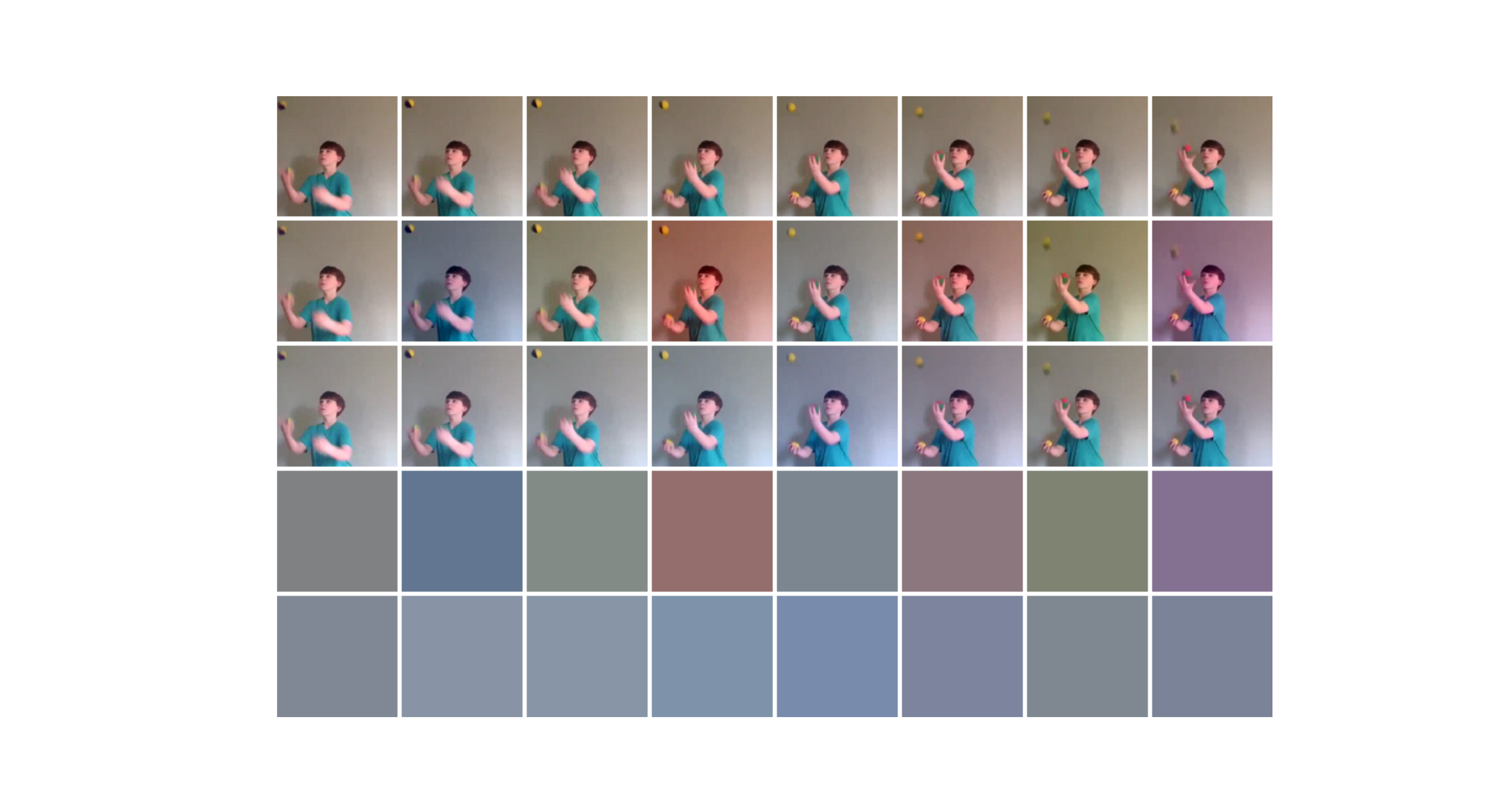}}
  {\caption{Illustration of the trade-off between thickness and roughness in a single video attack as described in Section \ref{thk_vs_rgh}.}
  \label{fig:juggling_balls_beta_1_compare}}
\end{figure}
\subsubsection{Thickness Vs. Roughness}\label{thk_vs_rgh}
In order to illustrate the trade-off between $\beta_{1}$ and $\beta_{2}$ under single video attacks, we have selected a video sample (kinetics test set) on which we developed two different flickering attacks by solving Equation (\ref{eq:opt_eq_gen}) (separately) under the single video attack settings $(N=1)$.
As described in Section \ref{Regularization terms}, the $\beta_{j}$'s coefficients control the importance of each regularization term, where $\beta_{1}$ associated with the term that forces the perturbation to be as small as possible in gray-level over the three color channels and $\beta_{2}$ associated with purely temporal terms (norms of the first and second temporal derivatives) forcing the perturbation to be temporally-smooth as possible.
The first perturbation developed with $\beta_{1}=1$ and $\beta_{2}=0$, minimize the thickness while leaving the roughness unconstrained. The second perturbation developed with $\beta_{1}=0$ and $\beta_{2}=1$, minimize the roughness while leaving the thickness unconstrained.
Both of these perturbations cause misclassifications on the I3D model.
In order to visualize the difference between these perturbations, we deliberately picked a difficult example to attack which that requires large thickness and roughness.
In Figure \ref{fig:juggling_balls_beta_1_compare} we plot both attacks in order to visualize the difference between the two cases. Each  row  combined 8 consecutive frames (out of 90 frames). In the first row, the original (clean) video sample from the \textit{``juggling balls"} category. In the second row, the adversarial (misclassified) video we developed with $\beta_{1}=1$ and $\beta_{2}=0$ (minimizing thickness). In the third row the adversarial video with $\beta_{1}=0$ and $\beta_{2}=1$ (minimizing roughness).
In the fourth row we plot the flickering perturbations with $\beta_{1}=1$, $\beta_{2}=0$ reaching a thickness of $2.97\%$ and roughness of $4.84\%$.
In the fifth row we plot the flickering perturbations with
$\beta_{1}=0$, $\beta_{2}=1$ reaching a thickness of $7.45\%$ and roughness of $2.20\%$.
As expected, the perturbation with the minimized roughness (last row) is smoother than the one without the temporal constrain (fourth row). Furthermore, even though the thickness of temporal constrained perturbation is much higher (7.45\% compare to 2.97\%) the adversarial perturbation is less noticeable to the human observer than the one with the smaller thickness.
Video examples of the discussed attacks can be found here\textsuperscript{\ref{youtube_videos}} under \textit{``juggling balls"}.
\subsection{Adversarial Attack Generalization}\label{Adversarial Attack Generalization}
Unlike single video attack, where the flickering perturbation $\delta$ was video-specific, a generalized (or universal) flickering attack is a single perturbation that fools our targeted model with high probability for all videos (from any class or a specific class).
In order to obtain a universal adversarial perturbation across videos we solve the optimization problem in Equation (\ref{eq:opt_eq_gen}) with some attack-specific modifications as described in the following sections.
\subsubsection{Class generalization: Untargeted Attack}\label{class_gen}
Adversarial attacks on a single video have limited applicability in the real world. In this section we generalize the attack to cause misclassification to all videos from a specific class with a single generalized adversarial perturbation $\delta$.
Our experiments conducted on 100 (randomly picked) out of 400 kinetics classes s.t.
for each class (separately) we developed its own $\delta$ by solving the optimization problem in Equation (\ref{eq:opt_eq_gen}), where $\{X_{n}\}_{n=1}^{N}$ is the relevant class training set split.
After developing the class generalization $\delta$  we evaluate its fooling ratio performance, thickness and roughness as defined in Section \ref{Metric} on the relevant class evaluation split.
The \textit{Single Class} entry in Table \ref{table:result} shows the statistics of average and standard deviation (across 100 different $\delta$'s) of the fooling ratio, thickness and roughness. Showing that when applying this attack, on average $90.2\%$ of the videos from each class were misclassified.
It is obvious that generalization produces perturbation with larger thickness and roughness.

\subsubsection{Universal Untargeted Attack}\label{univerasl_attack}
We take one more step toward real world implementation of the flickering attack by devising a single universal perturbation that will attack videos from any class.
Constructing such flickering attacks is not trivial due to the small number of trainable parameters (${T\times C}$ or $270$ in I3D) and in particular that they are independent of image dimensions.
Similarly to the previous section, we developed single $\delta$ by solving the optimization problem in Equation (\ref{eq:opt_eq_gen}), where $\{X_{n}\}_{n=1}^{N}$ is
the training set defined as the entire evaluation-split ($~20K$ videos) of the Kinetics-400. Once the universal $\delta$ was computed, we evaluated its fooling ratio performance, thickness and roughness on a random sub-sample of $5K$ videos from the kinetics test-split. As can be seen in \textit{Universal Class} entry in Table \ref{table:result}, our universal attack reaches a $93\%$ fooling ratio.
One might implement the universal flickering attack as a class-targeted attack using the presented method. In this case, the selected class may affect the efficiency of the adversarial perturbation.
\subsection{Time Invariance}\label{time_invariance}
Practical implementation of adversarial attacks on video classifiers can not be subjected to prior knowledge regarding the frame numbering or temporal synchronization of the attacked video. In this section we present a time-invariant adversarial attack that can be applied to the recorded scene without assuming that the perturbation of each frame is applied at the right time. Once this time-invariant attack is projected to the scene in a cyclic manner, regardless of the frame arbitrarily-selected as first, the adversarial pattern would prove to be effective.  
Similar to the generalized adversarial attacks described in previous subsections, a random shift between the perturbation and the model input was applied during training. The adversarial perturbation in Equation (\ref{eq:opt_eq_gen}) modified by adding the $Roll$ operator defined in Equation (\ref{eq:roll})  s.t.~$F_{\theta}(X_{n}+Roll(\delta, \tau))$ for randomly sampled $\tau \in \{1,2, \cdots, T\}$ in each iteration and on each video during training and evaluation. This time invariance generalization of universal adversarial flickering attack reaches $83\%$ fooling ratio, which is luckily a small price to pay in order to approach real-world implementability.
\section{Additional models, baseline comparisons and transferability}
In order to demonstrate the effectiveness of the flickering adversarial attack (universal in particular) we applied selected attacks to other relevant models and compared between the proposed universal flickering attack to other baseline attacks (Section \ref{Baseline comparison}) and validate that our attack is indeed transferable \cite{Szegedy14} across models (Section \ref{Transferability}).
\subsection{Targeted Models}
Similar to the previous experiment we follow the white-box setting. In the following experiments we apply our attack on three different models \textbf{MC3, R3D, R(2+1)D} (pre-trained on the Kinetics Dataset) from \cite{tran2018closer} which discuss several forms of spatiotemporal convolutions and study their effects on action recognition.
All three model are based on 18 layers ResNet architecture \cite{he2016deep}, accepting spatial and temporal dimensions of: $T=16, H=112,W=112,C=3$.
Implementation details can be found in this paper's supplementary material.
\subsection{Baseline comparison}\label{Baseline comparison}
Following the introduction of the first flickering attack against video action recognition models, a baseline comparison of the effectiveness of the universal attack is presented against several types of random flickering perturbations.
We developed a universal flickering perturbation $\delta^{F}$ on model $F$ (I3D, R(2+1)D, etc.) with respect to the Kinetics Dataset by solving the optimization problem defined by Equation (\ref{eq:opt_eq_gen}).
Following Equation (\ref{eq:opt_eq_gen}) we constrained the $\ell_{\infty}$ norm of $\delta^{F}$ by clipping s.t.  $\left\|\delta^{F}\right\|_{\infty}=\max|\delta^{F}|\leq \zeta$ for some $\zeta$.

In order to evaluate the Fooling ratio of any $\delta$ (and in particular $\delta^{F}$) on some model $F$ we define the evaluation set $\mathbb{X}=\{X_{i}\}_{i=1}^{M}$ where $X_{i} =[x_{1}^{i}, x_{2}^{i},..,x_{T}^{i}]$ is $i^{th}$ evaluation video consisting of $T$ consecutive frames. On top of $\mathbb{X}$ we define the adversarial evaluation set  $\hat{\mathbb{X}}_{\delta}=\{\hat{X_{i}}\}_{i=1}^{M}$ where, $\hat{X_{i}} =[x_{1}^{i}+\delta, x_{2}^{i}+\delta,..,x_{T}^{i}+\delta]$ for all $i$. Therefore, the fooling ratio is calculated by evaluating $F$ on $\hat{\mathbb{X}}_{\delta}$.
In the following experiments we use the same evaluation set $\mathbb{X}$.

Given a flickering universal adversarial perturbation $\delta^{F}$ developed on model $F$, we define the following random flickering attacks:\\
$\delta_{U}^{F} \sim \mathcal{U}(\min\delta^{F} ,\max\delta^{F} )$: Random variable uniformly distributed between the minimal and maximal values of $\delta^{F} $.\\
$\delta_{MinMax}^{F} $: Each element is drawn from the set  $\{\min\delta^{F} ,\max\delta^{F} \}$ with equal probability.\\
$\delta_{shuffle}^{F} $: A random shuffle of $\delta^{F}$ along the frames and color channels.
Table \ref{table:ranodm_vs_vs_trns_ours_result} shows the results of our experiments where each experiment (different $\ell_{\infty}[\%]$) was performed as follows:
\begin{enumerate}
    \item For given $\zeta$ we developing $\delta^{F}$ for each one of our four attacked models: I3D, R(2+1)D, R3D and MC3.
    \item For each $\delta^{F}$ we developed  $\delta_{U}^{F}, \delta_{MinMax}^{F}, \delta_{shuffle}^{F}$ as described earlier.
    \item On each model $F$ we evaluate the fooling ratio of the following perturbation: Random flickering ($\delta_{U}^{F}, \delta_{MinMax}^{F}, \delta_{shuffle}^{F}$), universal flickering developed upon other models and universal flickering $\delta^{F}$.
\end{enumerate}
In our experiments the $\ell_{\infty}[\%]$ norm of $\delta$ is represented as the percentage of the allowed pixel intensity range ($V_{max}$-$V_{min}$). e.g., if $V_{max}=1$, $V_{min}=-1$ and $\ell_{\infty}[\%]=10$ than $\zeta=0.2$. 
In order to obtain statistical attributes we performed the experiments by re-perturbing the random generated $\delta$'s ($\delta_{U}^{F}, \delta_{MinMax}^{F}, \delta_{shuffle}^{F}$).
As shown in Table \ref{table:ranodm_vs_vs_trns_ours_result} we performed the experiments over several values of $\ell_{\infty}[\%]$: 5, 10, 15 and 20. The columns (with models names) represent the attacked model, while the rows represent the type of flickering attacks.
Random flickering attacks are located at the first 3 rows of each experiment, followed by the universal flickering attack trained upon other models (except I3D)\footnote{\label{I3D_trns} The transferabilty between I3D to the other models (and vice versa) were not evaluated because the input of the models is not compatible.}-- marked with (trns).
The universal flickering attack (ours) is located at the last row of each experiment.
Each cell holds the fooling ratio result (average and standard deviation in the case of random generated perturbations) when evaluating the model on the data with the relevant attack.
As can be seen, the universal flickering attack demonstrates superiority across all four models, over the transferable attacks and the random flickering attacks.
In addition to Table \ref{table:ranodm_vs_vs_trns_ours_result}, additional analysis is presented in the supplementary material. 
\begin{table}
\caption{Baseline comparison of the universal flickering attack to several types of random flickering attacks and transferability across different models.}
\label{table:ranodm_vs_vs_trns_ours_result}
\begin{center}
\resizebox{1.\columnwidth}{!}{
\begin{tabular}{|c|l|c|c|c|c|}
\hline
$\ell_{\infty}[\%]$ &Attack \textbackslash Model      &I3D              &R(2+1)D          &R3D                &MC3 \\ 
\midrule
\multirow{7}{*}{5}
                                &Random Uniform      &8.4$\pm$ 0.6\%   &4.9$\pm$ 0.8\%  &8.3$\pm$ 1.8\%   &11.0$\pm$ 1.9\% \\
                                &Random MinMax       &12.2$\pm$ 0.7\%  &9.0$\pm$ 2.3\%  &15.8$\pm$ 3.5\%  &17.4$\pm$ 3.8\% \\
                                &Filckering shuffle  &11.9$\pm$ 0.6\%  &9.4$\pm$ 1.7\%  &16.4$\pm$ 3.3\%  &16.5$\pm$ 2.5\% \\
\cmidrule{2-6}
                                &R(2+1)D (trns)           &-                &-           &27.6\%  &18.4\%\\
                                &R3D     (trns)           &-           &14.9\%  &-                &24.0\% \\
                                &MC3     (trns)           &-           &12.3\%  &31.4\%   &-               \\
\cmidrule{2-6}
                                &\textbf{Filckering} &\textbf{26.2\%}  &\textbf{23.3\%}  &\textbf{34.3\%}   &\textbf{41.3\%}\\
\midrule
\multirow{7}{*}{10}
                                &Random Uniform      &14.2$\pm$ 1.2\%  &10.7$\pm$ 3.3\%  &20.2$\pm$ 5.3\%  &17.9$\pm$ 3.1\% \\
                                &Random MinMax       &23.6$\pm$ 2.4\%  &19.2$\pm$ 4.8\%  &36.7$\pm$ 6.3\%  &30.0$\pm$ 3.7\% \\
                                &Filckering shuffle  &22.9$\pm$ 2.1\%  &18.3$\pm$ 5.5\%  &31.9$\pm$ 7.2\%  &25.9$\pm$ 3.7\% \\
\cmidrule{2-6}
                                &R(2+1)D (trns)           &-                &-           &52.7\%  &38.4\%\\
                                &R3D     (trns)           &-           &30.6\%  &-                &35.6\% \\
                                &MC3     (trns)           &-           &25.9\%  &50.5\%   &-               \\
\cmidrule{2-6}
                                &\textbf{Filckering} &\textbf{58.4\%}  &\textbf{47.2\%}  &\textbf{70.4\%}   &\textbf{55.3\%}\\
\midrule
\multirow{7}{*}{15}
                                &Random Uniform     &20.3$\pm$ 2.1\%  &16.0$\pm$ 4.7\%  &26.2$\pm$ 4.7\%  &24.2$\pm$ 1.8\% \\
                                &Random MinMax      &34.2$\pm$ 3.1\%  &28.1$\pm$ 7.9\%  &48.6$\pm$ 7.4\%  &36.4$\pm$ 4.9\% \\
                                &Filckering shuffle &29.3$\pm$ 3.1\%  &28.7$\pm$ 5.0\%  &44.6$\pm$ 8.7\%  &35.3$\pm$ 2.8\% \\
\cmidrule{2-6}
                                &R(2+1)D (trns)           &-                &-          &64.4\%  &48.4\%\\
                                &R3D     (trns)            &-         &39.5\%  &-                &50.7\% \\
                                &MC3     (trns)            &-         &40.7\%  &66.1\%   &-               \\
\cmidrule{2-6}
                                &\textbf{Filckering} &\textbf{78.1\%}  &\textbf{62.7\%}  &\textbf{83.4\%}   &\textbf{73.3\%}\\
\midrule
\multirow{7}{*}{20} 
                                &Random Uniform      &32.1$\pm$ 3.1\%  &22.2$\pm$ 5.7\%  &37.1$\pm$ 4.0\%  &30.0$\pm$ 4.5\%\\
                                &Random MinMax       &48.0$\pm$ 4.5\%  &42.0$\pm$ 3.0\%  &54.6$\pm$ 11.0\% &44.0$\pm$ 5.0\% \\
                                &Filckering shuffle  &42.0$\pm$ 3.6\%  &39.0$\pm$ 8.0\%  &57.6$\pm$ 6.4\%  &47.1$\pm$ 4.7\%\\
\cmidrule{2-6}
                                &R(2+1)D (trns)            &-                &-          &74.6\%  &59.2\%\\
                                &R3D     (trns)            &-          &58.5\%  &-                &60.7\% \\
                                &MC3     (trns)            &-          &55.8\%  &70.4\%   &-               \\
\cmidrule{2-6}
                                &\textbf{Filckering} &\textbf{93.0\%}  &\textbf{79.0\%}  &\textbf{90.3\%}   &\textbf{77.1\%}\\
\bottomrule
\end{tabular}
 }
\end{center}
\end{table}
\subsection{Transferability across Models}\label{Transferability}
 Transferability \cite{Szegedy14} is defined as the ability of an attack to influence a model which was unknown to the attacker when developing the attack. We examined the transferability of the flickering attack on different models of the same input type. As seen in Table \ref{table:ranodm_vs_vs_trns_ours_result}, for each $\ell_{\infty}[\%]$ we evaluate the fooling ratio of attacks that was trained on different models (trns). The high effectiveness of the attack applied across models indicates that our attack is transferable between these different models, e.g., attack that was developed on R(2+1)D with $\ell_{\infty}[\%]=20$ achieved $74.6\%$ fooling ratio when applied on R3D model compared to $90.3\%$.  



\section{Over-the-Air Real world demonstration}
The main advantage of the flickering attack, unlike the majority of adversarial attacks in published papers, is its real-world implementability.
In this section we demonstrate, for the first time, the flickering attack in a real world scenario.
We used an RGB led light bulb and controlled it through Wifi connection. Through this connection we were able to control the red, green and blue illumination values separately, and create almost any of the previously developed adversarial RGB patterns introduced in this paper (Figure \ref{fig:overview_digital_and_ota} depicts the modeling of our digital domain attack in the real-world).  As in \cite{li2019adversarial,eykholt2018robust}, we have applied several constraints for better efficiency of the adversarial manipulations in real-world, such as temporal invariance (Section \ref{time_invariance}) and increased smoothness to address the finite rise (or fall) time of the RGB bulb (Section \ref{Roughness_regularization}). Because the adversarial patterns presented here have positive and negative amplitude perturbations, the baseline illumination of the scenario was set to around half of the possible maximum illumination of the bulb. A chromatic calibration of the RGB intensities was performed in order to mitigate the difference of the RGB illumination of the light bulb and RGB responsivity of the camera, which was obviously not the same and also included channel chromatic crosstalk.
The desired scenario for the demonstration of the attack includes a video camera streaming a video filmed in a room with a Wifi-controlled RGB light bulb. A computer sends over Wifi the adversarial RGB pattern to the bulb. A figure performs actions in front of the camera. Implementation and hardware details can be found in the supplementary material.
We demonstrate our over-the-air attack in two different ways, scene-based and universal flickering attack.

\begin{figure}[ht]
  \centering
    \includegraphics[width=\linewidth,trim=1.5cm 0cm 3.5cm 1.2cm,clip]{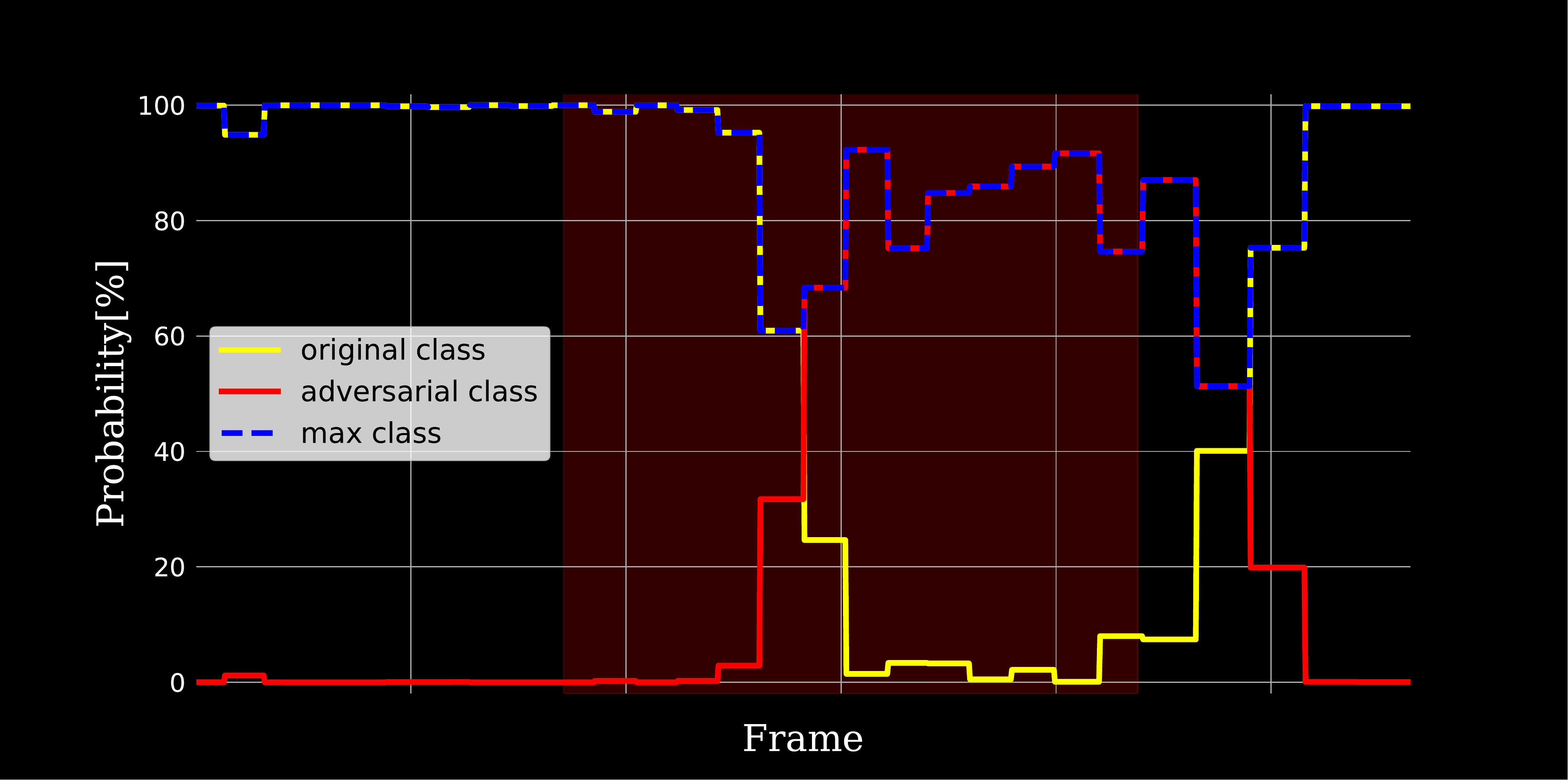}
    \caption{Example of our Over-the-Air scene based attack.  The plot was taken from the \textit{``ironing"} video example\textsuperscript{\ref{ota_scene_youtube_videos}}.}
    \label{fig:ota_ironing_scene_attack}
\end{figure}
\subsection{Over-the-Air Scene-based Flickering Attack}
In this attack, we assume prior knowledge of the scene and the action.
Therefore, similar to a single video attack (Section \ref{sec:single_vid_attack}) we will develop a scene dedicated attack.
In this approach we record a clean video (without perturbation) of the scene we would like to attack.
For the clean recording we develop a time-invariant digital attack as described in the paper.
Once we have the digital attack, we transmit it to a ``similar" scene (as described in the supplementary material) in an over-the-air approach.
Video examples of our scene based over-the-air adversarial attack can be found here\textsuperscript{\ref{ota_scene_youtube_videos}}. 
Figure \ref{fig:ota_ironing_scene_attack} shows the probability of a real example of our scene based over-the-air attack of the \textit{``ironing"} action, where the x-axis (Frame) represents prediction time step and the y-axis (Probability) represents the output probability of the I3D model for several selected classes.
The area shaded in red represents the period of time the scene was attacked.
As described in the legend, the yellow graph is the true class (\textit{``ironing"}) probability, the red graph is the adversarial class (\textit{``drawing"}) probability and the dashed blue graph represents the probability of the most probable class the classifier predicts each frame.
It can be seen that when the scene is not attacked (outside the red area) the model predicts correctly the action being performed (dashed blue and yellow graphs overlap). Once the scene is attacked, the true class is suppressed and the adversarial class is amplified.
At the beginning (end) of the attack, it can be seen that there is a delay from the moment the attack begins (ends) until the model responds to the change due to the time required (90 frames) to fill the classifier's frame buffer and perform the prediction.


\subsection{Over-the-Air Universal Flickering Attack}
This section deals with the case where we do not have any prior knowledge regarding the scene and action we wish to attack.
Therefore, we would like to develop a universal attack that will generalize to any scene or action.
In this approach, we will use a universal time-invariant attack as described in the paper.
Once we have the digital attack, we transmit it to the scene in an over-the-air approach.
Video examples of our universal over-the-air attack can be found here\textsuperscript{\ref{ota_youtube_videos}}.
Since our approach is real-world applicable, and thus we require universality and time-invariability perturbation (no need to synchronize the video with the transmitted perturbation), the pattern is visible to the human observer.
\section{Conclusions and future
work}\label{Conclusions and Future Work}
The flickering adversarial attack was presented, for the first time, for several models and scenarios summarized in Tables \ref{table:result}, \ref{table:ranodm_vs_vs_trns_ours_result}. Furthermore, this attack was demonstrated in the real world for the first time. The flickering attack has several benefits, such as the relative imperceptability to the human observer in some cases, achieved by small and smooth perturbations as can be seen in the videos we have posted\textsuperscript{\ref{youtube_videos}}.
The flickering attack was generalized to be universal, demonstrating superiority over random flickering attacks on several models.
In addition, the flickering attack has demonstrated the ability to transfer between different models.
The flickering adversarial attack is probably the most applicable real-world attack amongst any video adversarial perturbation this far, as was shown\textsuperscript{\ref{ota_youtube_videos},\ref{ota_scene_youtube_videos}}.
Thanks to the simplicity and uniformity of the perturbation across the frame which can be achieved by subtle lighting changes to the scene by illumination changes. All of these properties make this attack implementable in real-world scenarios. 

In extreme cases where generalization causes the pattern to be thick enough to be noticed by human observers, the usage of such perturbations can be relevant for non-man-in-the-loop systems or cases where the human observer will see image-flickering without associating this flickering with an adversarial attack. 
In the future, we may expand the current approach to develop defensive mechanisms against adversarial attacks of video classifier networks.
\newpage

{\small
\bibliographystyle{ieee_fullname}
\bibliography{egbib}
}
\clearpage

\begin{appendices}
\let\clearpage\relax
\section{Modified Adversarial loss function}\label{appendix:Modified Adversarial loss function}
For achieving a more stable convergence, we used a loss mechanism similar to the loss presented by \cite{CarliniW16a}, with a small modification, which smoothly reaches the adversarial goal only to the desired extent, leaving space for other regularization terms. For untargeted attack:
\begin{equation}\label{eq:adv_loss_supp}
\ell(y,t) =
\max \left(0,  \min\left(\frac{1}{m} \ell_{m}(y,t)^2,
\ell_{m}(y,t)\right) \right)
\end{equation}
\begin{equation}\label{eq:sub_adv_loss_supp}
\ell_{m}(y,t) = y_{t}- \max_{i\neq t}(y_{i}) +m.
\end{equation}
$m>0$ is the desired margin of the original class probability below the adversarial class probability.
When loss values are within the desired margin, the quadratic loss term relaxes the relatively steep gradients and momentum of the optimizer, and the difference between the first and second class probabilities approach the desired margin $m$. When the loss starts rising, the quadratic term gently maintains the desired difference between these two classes, therefore preventing overshoot effects.
In order to apply the suggested mechanism on targeted attack, the loss term changed to $\ell_{m}(y,t) =\max_{i\neq t}(y_{i}) -y_{t}  +m$, while this time, $t$ is the targeted adversarial class.

In some cases it would be beneficial to follow \cite{CarliniW16a} and use the logits instead of the probabilities for calculating the loss. We suggest adapting this method partially by keeping the desired margin in probability space, normalized at each iteration accordingly, for margin defined in logit space may be less intuitive as a regularization term.

\begin{figure}[ht]
    \includegraphics[width=\columnwidth]{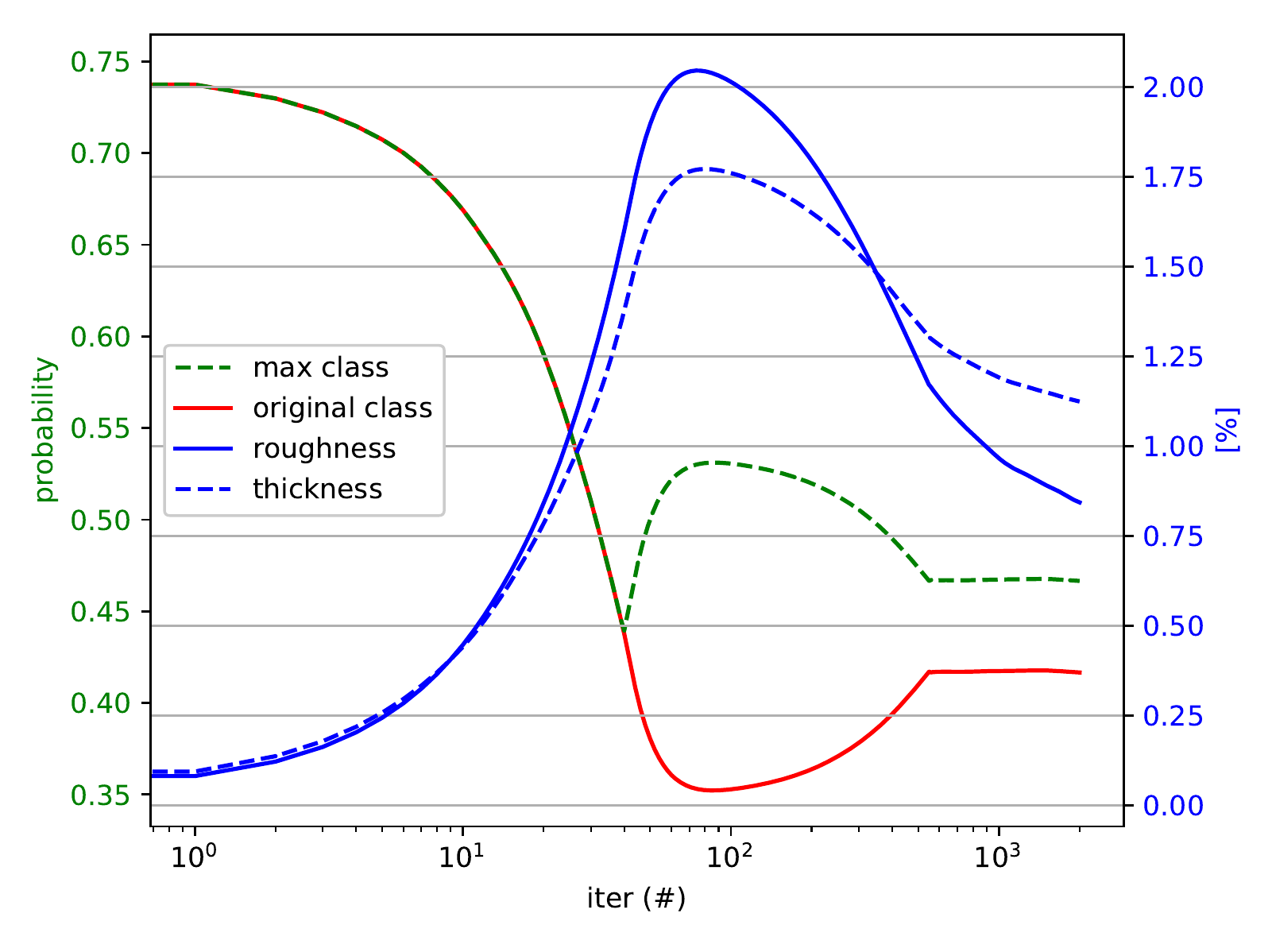}
    \caption{Learning process of the modified loss mechanism. Probabilities (green and red lines) corresponds to the left y-scale. Roughness and thickness (blue lines) are in percents from the full gray-level range of the image (right y-scale). \textit{original class} is the probability of the actual class of the unperturbed video. \textit{max class} is the probability of the most probable class as the classifier predicts.}
    \label{fig:porb_smoothness_thickness}
\end{figure}
\section{Implementation Details}
\subsection{Experiments on I3D}
Experiment codes are implemented in TensorFlow\footnote{\url{https://www.tensorflow.org/}} and based on I3D source code\footnote{\url{https://github.com/deepmind/kinetics-i3d}}. The code is executed on a server with four Nvidia Titan-X GPUs, Intel i7 processor and 128GB RAM. For optimization we adopt the ADAM \cite{kingma2014method} optimizer with learning rate of 1e-3 and with batch
size of 8 for the generalization section and 1 for a single video attack.
Except where explicitly stated $\beta_{1}=\beta_{2}=0.5$.
For single video attack and for generalization sections $\lambda=1$.
\subsection{Experiments on MC3, R3D, R(2+1)D}
Experiments code are implemented in PyTorch\footnote{\url{https://pytorch.org/}} and based on source code of computervision-recipes\footnote{\url{https://github.com/microsoft/computervision-recipes}} and torchvision\footnote{\url{https://github.com/pytorch/vision}}  package. 
Hardware, optimizer, batch size, $\beta_{1},\beta_{2}$ and $\lambda$ are the same as previously introduced for the I3D model.

\section{Single Video Attack}
\subsection{Convergence Process}\label{appendix:Convergence Process}
In order to demonstrate the convergence process we have attacked a single video. As can be seen, several trends regarding the trends can be observed (Figure \ref{fig:porb_smoothness_thickness}). At first, the adversarial perturbation rises in thickness and roughness. At iteration $40$ the top-probability class switches from the original to the adversarial class, which until now was not plotted, for this adversarial attack is untargeted. At that iteration, the adversarial loss is $m$. When the difference between the probability of the adversarial and original class is larger then $m$ the adversarial loss is zero and the regularization starts to be prominent, causing the thickness and roughness to decay. This change of trend occurs slightly after the adversarial class change due to the momentum of the Adam optimizer and remaining intrinsic gradients. At iteration $600$ the difference between the probability of the adversarial and original class is $m=0.05$, the quadratic loss term maintaining the desired difference between these classes while diminishing the thickness and roughness. The binary loss changes at the interface between adversarial success and failure  caused convergence issues, and the implementation of the quadratic term, as defined in Equation (\ref{eq:adv_loss_supp}) handled this issue.
\subsection{Thickness Vs. Roughness}\label{appendix:Thickness Vs. Roughness}
\begin{figure*}
  \centering
  \includegraphics[scale=0.6,trim=2.5cm 1cm 3cm 2cm,clip]{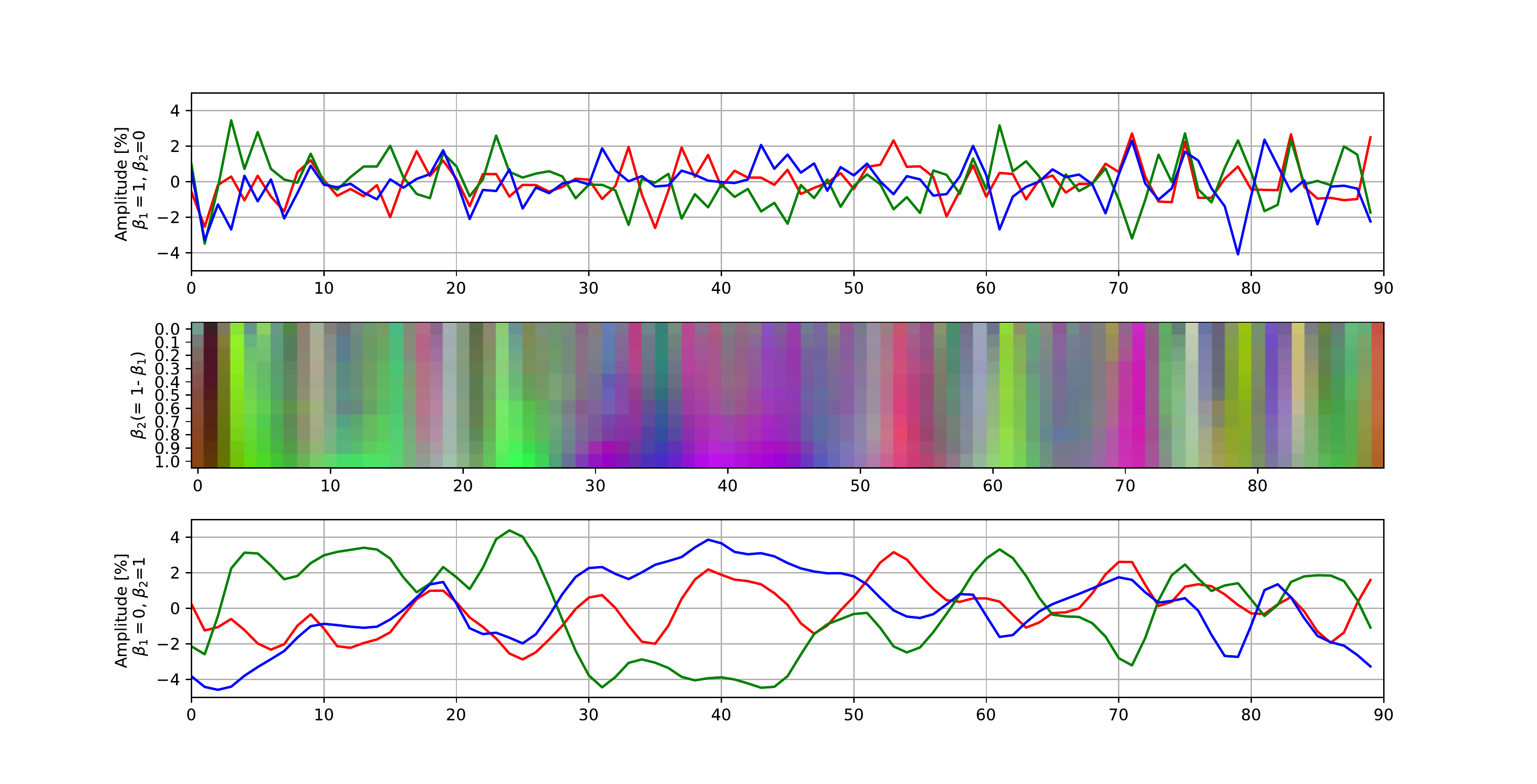}
  \caption{Top: The adversarial perturbation of the RGB channels (color represents relevant channel) as a function of the frame number at the case that $\beta_{1}$ = 1 and $\beta_{2}$ = 0 ($D_{1}$ minimization is preferred). Bottom: The adversarial perturbation of the RGB channels as a function of the frame number at the case that $\beta_{1}$ = 0 and $\beta_{2}$ = 1 ($D_{2}$ minimization is preferred). Top and bottom graphs are presented in percents from the full scale of the image. Middle: The gradual change of the adversarial pattern between the two extreme cases  where $\beta_{1}$ = 0 corresponds to the top graph and $\beta_{1}$ = 1 corresponds to the bottom graph. Color (stretched for visualization purposes) represents the RGB parameters of the adversarial pattern of each frame.}
  \label{fig:colormap}
\end{figure*}

In order to visualize the trade-off between $\beta_{1}$ and $\beta_{2}$ we plotted three graphs in Figure \ref{fig:colormap}. In top and bottom graphs we see the temporal amplitude of the adversarial perturbation of each frame and for each color channel, respectively. The extreme case (top) of minimizing only $D_{1}$ (given success of the untargeted adversarial attack) and leaving $D_{2}$ unconstrained ($\beta_{1}$ = 1, $\beta_{2}$ = 0). The signal of the RGB channels fluctuates strongly with a thickness value of 0.87\% and a roughness of 1.24\%. The other extreme case (bottom) is when $D_{2}$ is constrained and $D_{1}$ is not ($\beta_{1}$ = 0, $\beta_{2}$ = 1), leading to a thickness value of 1.66\% and a roughness value of 0.6\%. The central image displays all the gradual cases between the two extremities: $\beta_{1}$ goes from 1 to 0, and $\beta_{2}$ from 0 to 1 on the y-axis. The row denoted by $\beta_{2}$ = 0 corresponds to the upper graph and the row denoted by $\beta_{2}$ = 1 corresponds to the lower graph. Both $D_{1}$ and $D_{2}$ are very dominant in the received perturbation, as desired. Visualization of the path taken by our loss mechanisms at different $\beta_{1}$ and $\beta_{2}$ values can be found in the supplementary material.

\begin{figure*}
   \centering
   \includegraphics[width=\textwidth,trim=3cm 1cm 4cm 2cm,clip]{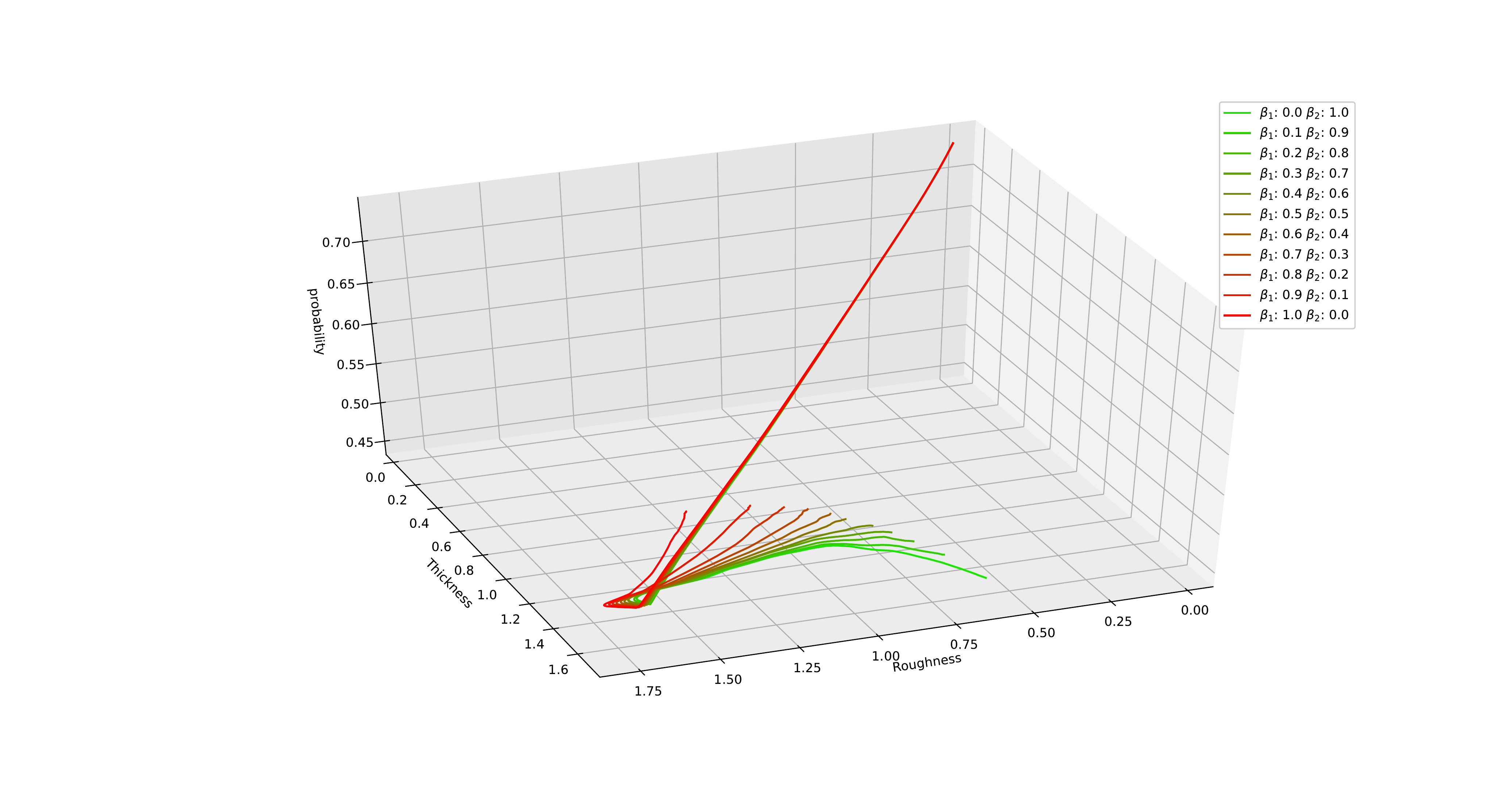}
   \caption{Convergence curve in probability-thickness-roughness space of an untargeted adversarial attack with different $\beta_{1}$ and $\beta_{2}$ parameters.}
   \label{fig:3D_paths}
\end{figure*}

Apart from the visualization experiments we showed, another experiment have been conducted in order to visualize the path taken by our loss mechanism at different $\beta_{1}$ and $\beta_{2}$. We have plotted a 3D representation in probability-thickness-roughness space for $10$ different experiments ($10$ different single video attack on the same video) with gradual change of $\beta_{1}$ and $\beta_{2}$ parameters.  
Figure \ref{fig:3D_paths} shows the probability of the most probable class at 10 different scenarios as described in the legend. One can see that at the beginning the maximal probability (original class) drops from the initial probability (upper section of the graph) on the same path for all of the described cases, until the adversarial perturbation takes hold of the top class. From there, the $\beta$'s parameters takes the lead. At this point, each different case is converging along a different path to a different location on the thickness-roughness plane. The user may choose the desired ratios for each specific application.

\section{Additional models, baseline comparison and transferability}
\subsection{Baseline comparison}
\begin{figure*}
\begin{adjustbox}{minipage=\linewidth,scale=1.}
\centering
\begin{subfigure}{0.5\columnwidth}
  \centering
  \includegraphics[width=\columnwidth,trim=0.7cm 0cm 1cm 0.7cm,clip]{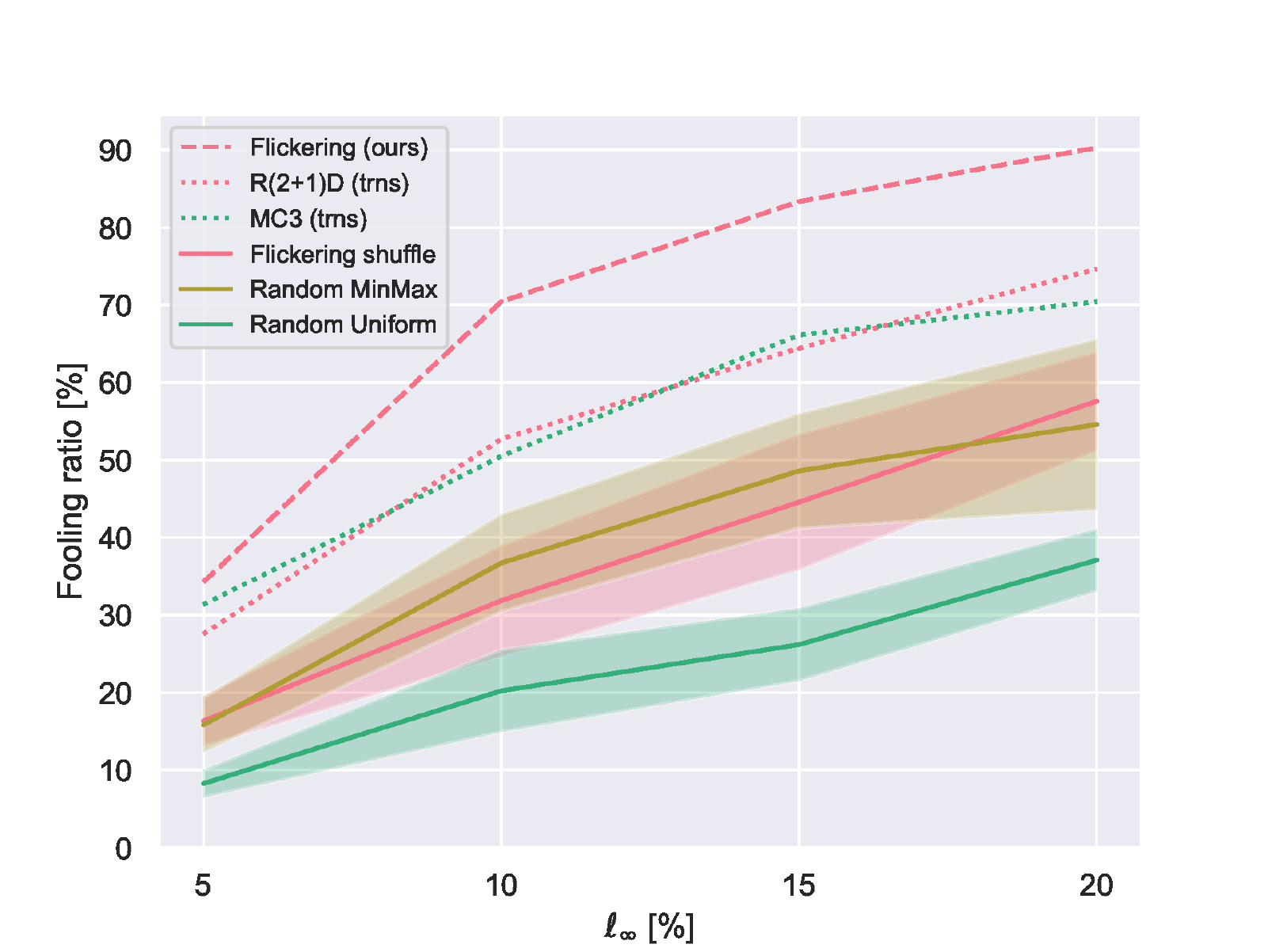}
  \caption{R3D}
  \label{fig:r3d_18_trns_rnd}
\end{subfigure}%
\begin{subfigure}{0.5\columnwidth}
  \centering
  \includegraphics[width=\columnwidth,trim=0.7cm 0cm 1cm 0.7cm,clip]{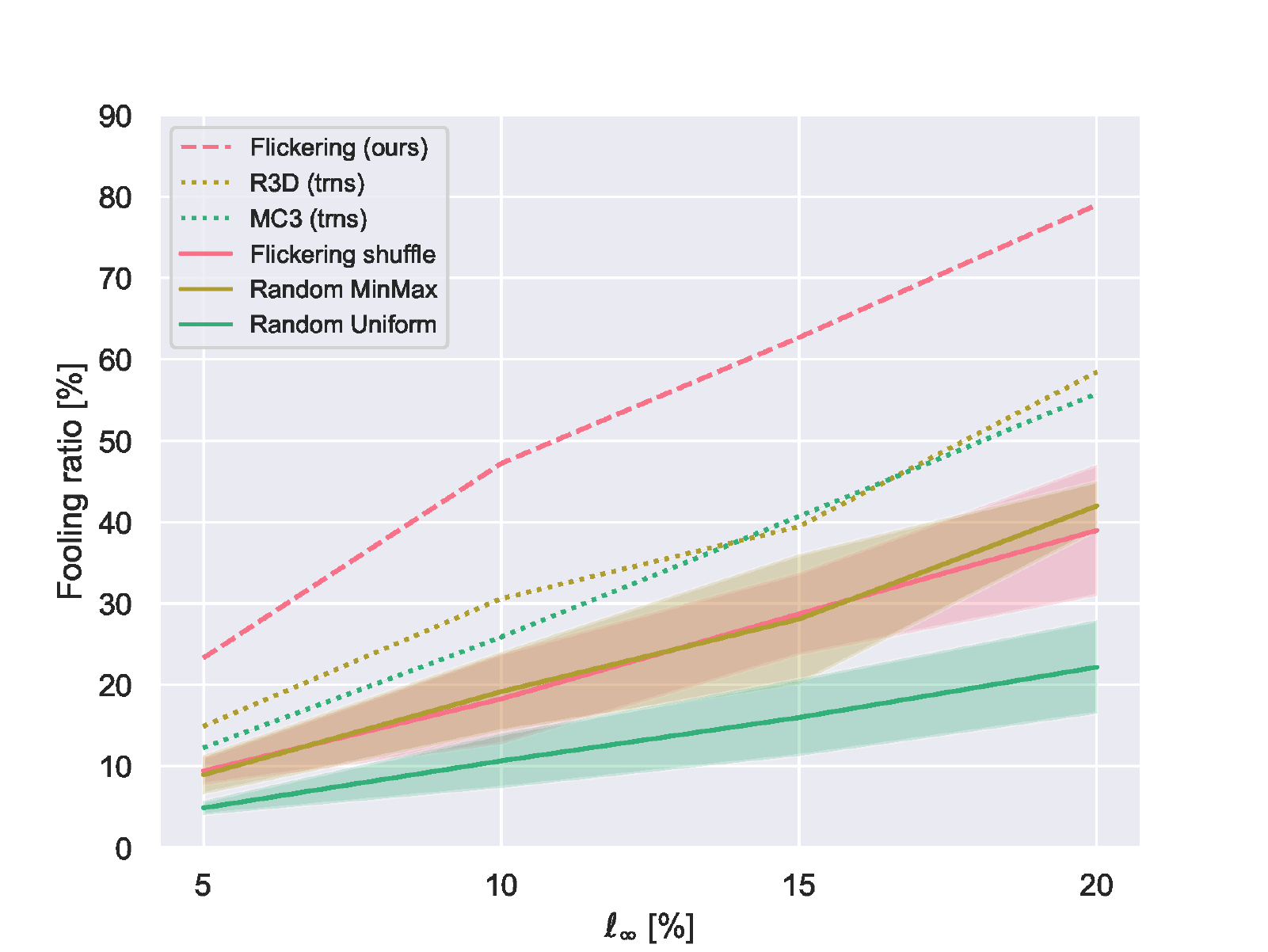}
  \caption{R(2+1)D}
  \label{fig:r2plus1d_18_trns_rnd}
\end{subfigure}
\begin{subfigure}{0.5\columnwidth}
  \centering
  \includegraphics[width=\columnwidth,trim=0.7cm 0cm 1cm 0.7cm,clip]{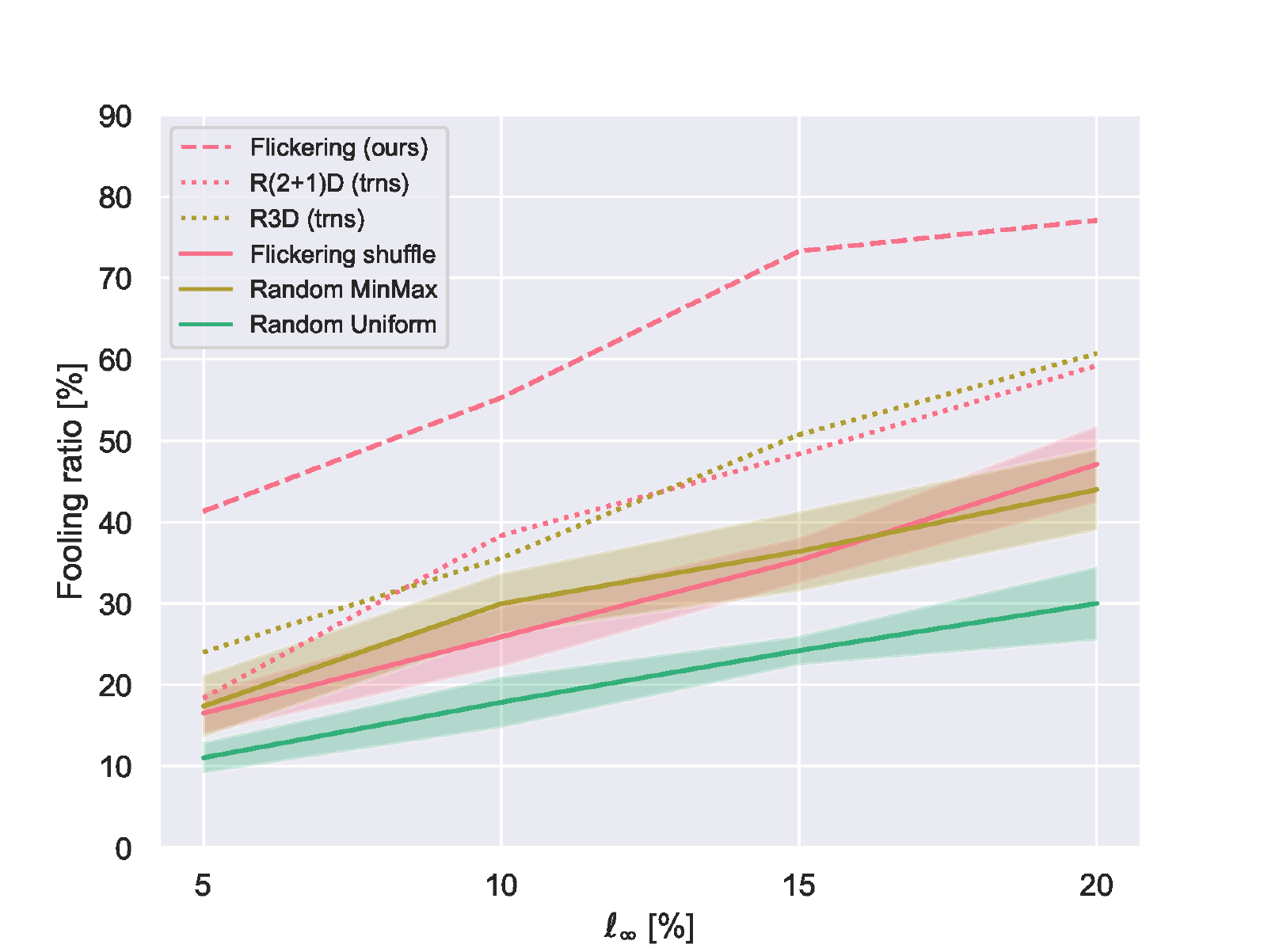}
  \caption{MC3}
  \label{fig:mc3_18_trns_rnd}
\end{subfigure}%
\begin{subfigure}{0.5\columnwidth}
  \centering
  \includegraphics[width=\columnwidth,trim=0.7cm 0cm 1cm 0.7cm,clip]{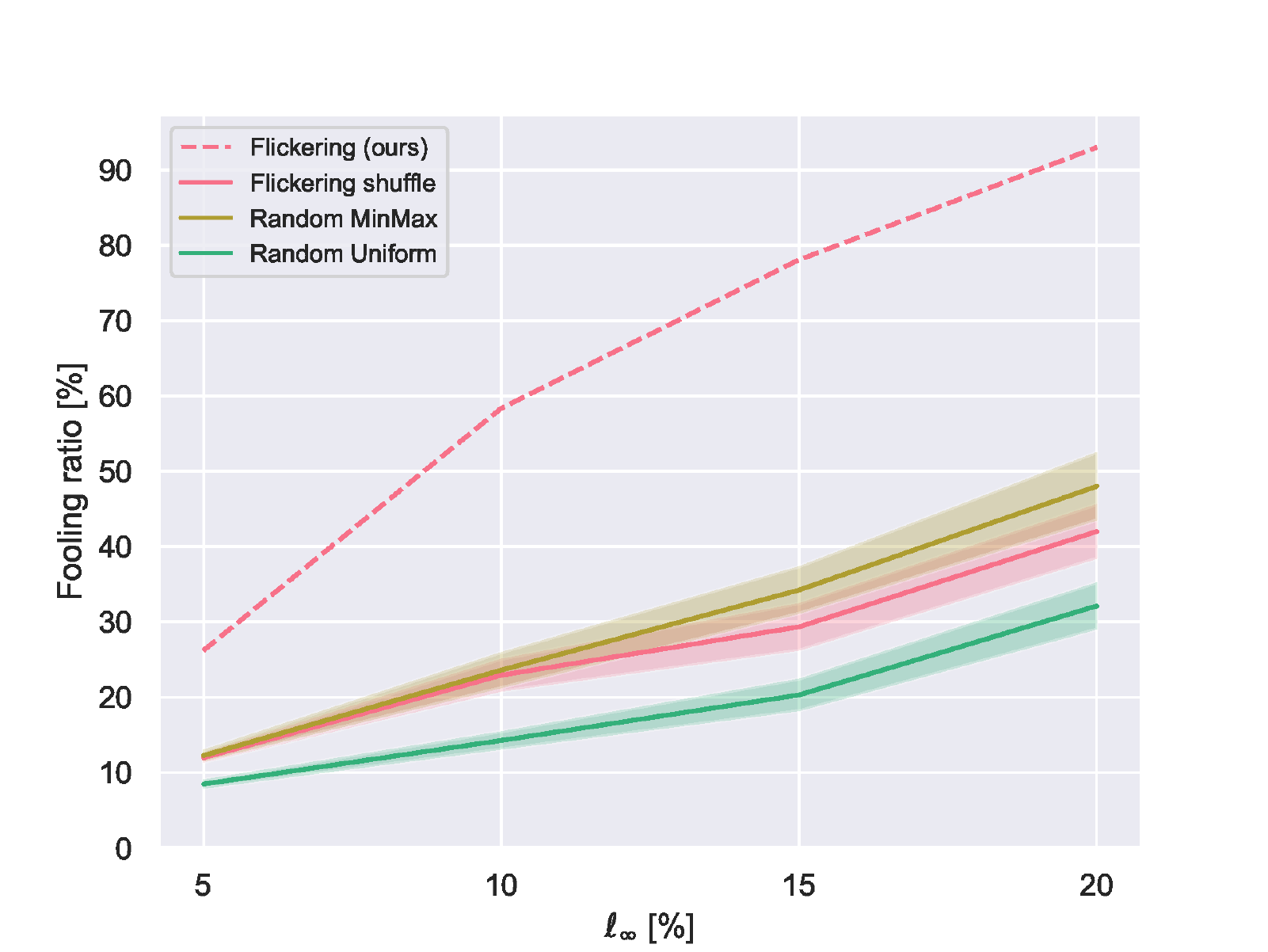}
  \caption{I3D}
  \label{fig:I3D_trns_rnd}
\end{subfigure}
\caption{Each one of the sub-figures shows the average fooling ratio of the attacked model (described in caption) with different perturbations as a function of $\ell_{\infty}[\%]$.
Each sub-figure combine three (two in I3D) main graph types, the dashed graph represent the Universal flickering perturbation developed on the attacked model ($\delta^{F}$), the dotted graphs represent the universal flickering attack developed upon other models (except for I3D) and the continues graphs represent the random generated flickering perturbations ($\delta_{U}^{F}, \delta_{MinMax}^{F}, \delta_{shuffle}^{F}$) where the shaded filled region is $\pm$ standard deviation around the average Fooling ratio.}
\label{fig:model_fr_rnd_trns_flickering}
\end{adjustbox}
\end{figure*}

In addition to the table presented in the paper,
we have analyzed our experiments from the attacked model perspective.
Each Sub-figure in Figure \ref{fig:model_fr_rnd_trns_flickering} shows the average fooling ratio of the attacked model (out of four) with different perturbation as function of $\ell_{\infty}[\%]$.
Each sub-figure combine three (two in I3D)\textsuperscript{\ref{I3D_trns}} main graph types, the dashed graph represent the universal flickering perturbation developed upon the attacked model ($\delta^{F}$), the dotted graphs represent the universal flickering attack developed upon other models (except for I3D) and the continues graphs represent the random generated flickering perturbation ($\delta_{U}^{F}, \delta_{MinMax}^{F}, \delta_{shuffle}^{F}$) where the shaded filled region is $\pm$ standard deviation around the average fooling ratio.
Several consistent trends can be observed in each one of the sub-figure and thus for each attacked model.
For each $\ell_{\infty}[\%]$ we can see that the fooling ratio order (high to low) is, first universal flickering attack, then the transferred universal flickering attack developed upon other models and finally, the random generated flickering perturbations.
\begin{figure*}
\begin{adjustbox}{minipage=\linewidth,scale=1.}
\centering
\begin{subfigure}{0.45\columnwidth}
  \centering
  \includegraphics[width=\columnwidth]{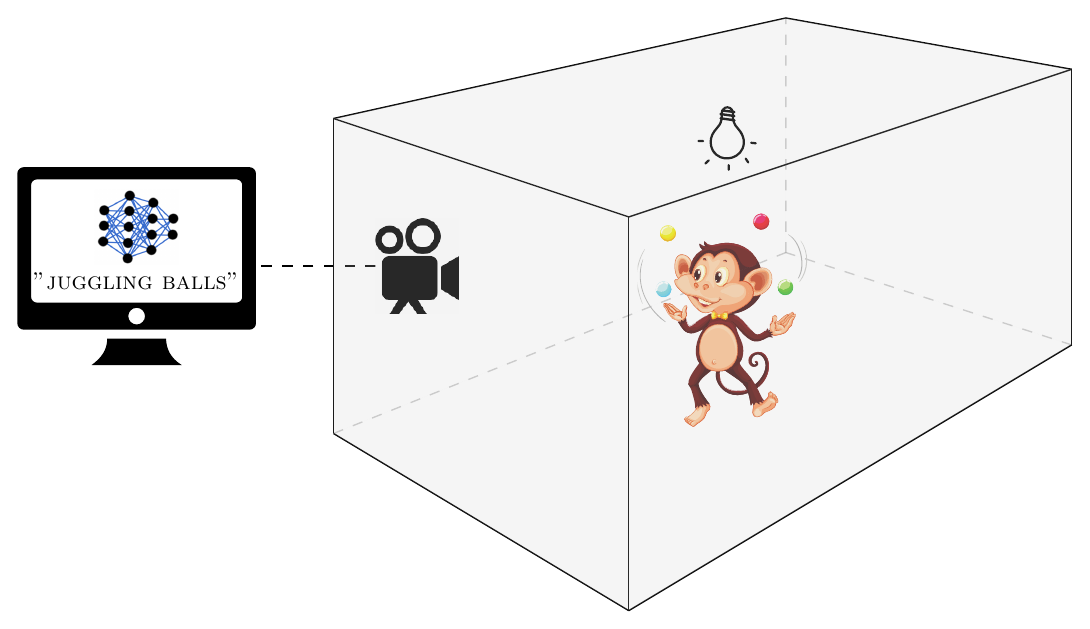}
  \caption{Without over-the-air attack, the action recognition network classify the action correctly as "juggling balls".}
  \label{fig:ota_room_attack_off}
\end{subfigure}%
\hspace{5mm}
\begin{subfigure}{0.45\columnwidth}
  \centering
  \includegraphics[width=\columnwidth]{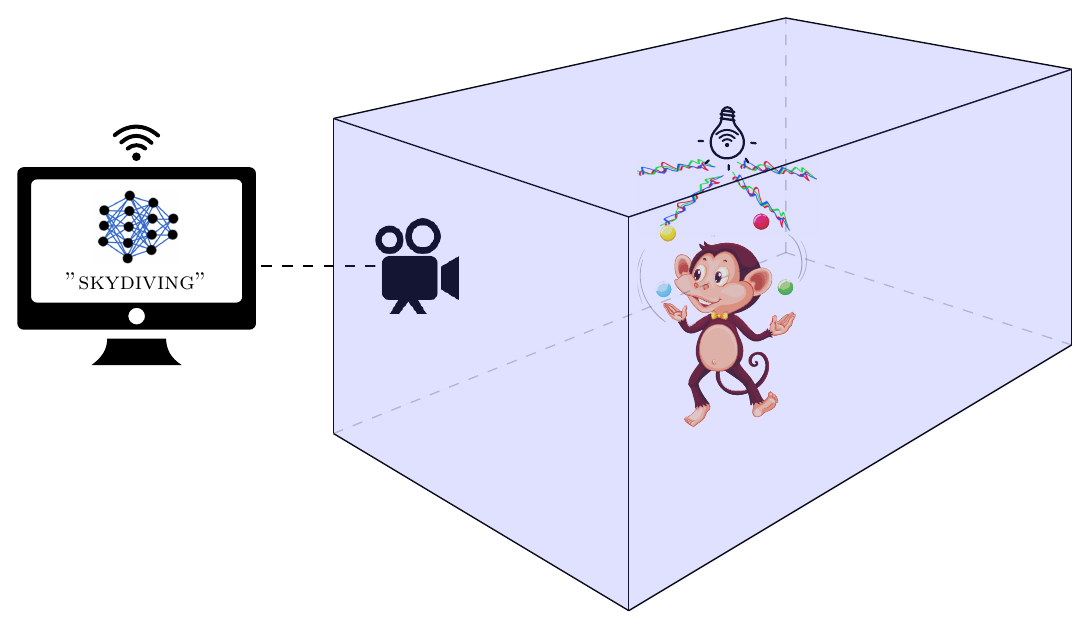}
  \caption{With over-the-air attack, the action recognition network classify the action incorrectly as "skydiving".}
  \label{fig:ota_room_attack_on}
\end{subfigure}

\caption{Room sketched of our over-the-air attack setup.}
\label{fig:ota_room}
\end{adjustbox}
\end{figure*}

\section{Over-the-Air Real world demonstration}
Our goal is to produce an adversarial universal flickering attack,
which will be implemented in the real world by an RGB led light bulb in a room, causing miss-classification.
The desired scenario for the demonstration of the attack includes a video camera streaming a video filmed in a room with a Wifi-controlled RGB led light bulb. A computer sends over Wifi the adversarial RGB pattern to the bulb. A figure performs actions in front of the camera. 
The hardware specifications are as follows:
\begin{itemize}
    \item \textbf{Video camera}: We used $1.3$ MPixel RGB camera streaming at 25 frames per second.
    \item \textbf{RGB led light bulb}: In order to applying the digitally developed (univrsel or scene based) perturbation to the scene, we use a RGB led light bulb\footnote{\url{https://www.mi.com/global/mi-led-smart-bulb-essential/specs}}, controlled over Wifi via Python  api\footnote{\url{https://yeelight.readthedocs.io/en/latest/}}, allowing to set RGB value at relatively high speed. 
    \item \textbf{Computer}: We use a computer to run the I3D action classifier on the streaming video input. The model input for prediction at time $t$ are all consecutive frames between $t-90$ to $t$ (as described in I3D experiments section). The model prediction frequency is set to $2$Hz (hardware performance limit).
    In addition, we use the computer in order to control the smart led bulb.
    \item \textbf{Acting figure}: Performs the actions we would like to classify and attack.  
\end{itemize}

Figure \ref{fig:ota_room} demonstrate our over-the-air attack setup, combining the hardware mentioned above.
Figure \ref{fig:ota_room_attack_off} demonstrate the state when the attack is off (no adversarial pattern is transmitted) and the video action recognition network correctly classify the action, while Figure \ref{fig:ota_room_attack_on} demonstrate the state when the attack is on (adversarial pattern is transmitted) and 
the video action recognition network incorrectly classify the action.

\begin{figure*}
\begin{adjustbox}{minipage=\linewidth,scale=1.}
\centering
\begin{subfigure}{0.45\columnwidth}
  \centering
  \includegraphics[width=\columnwidth,trim=0.7cm 0cm 1cm 0.7cm,clip]{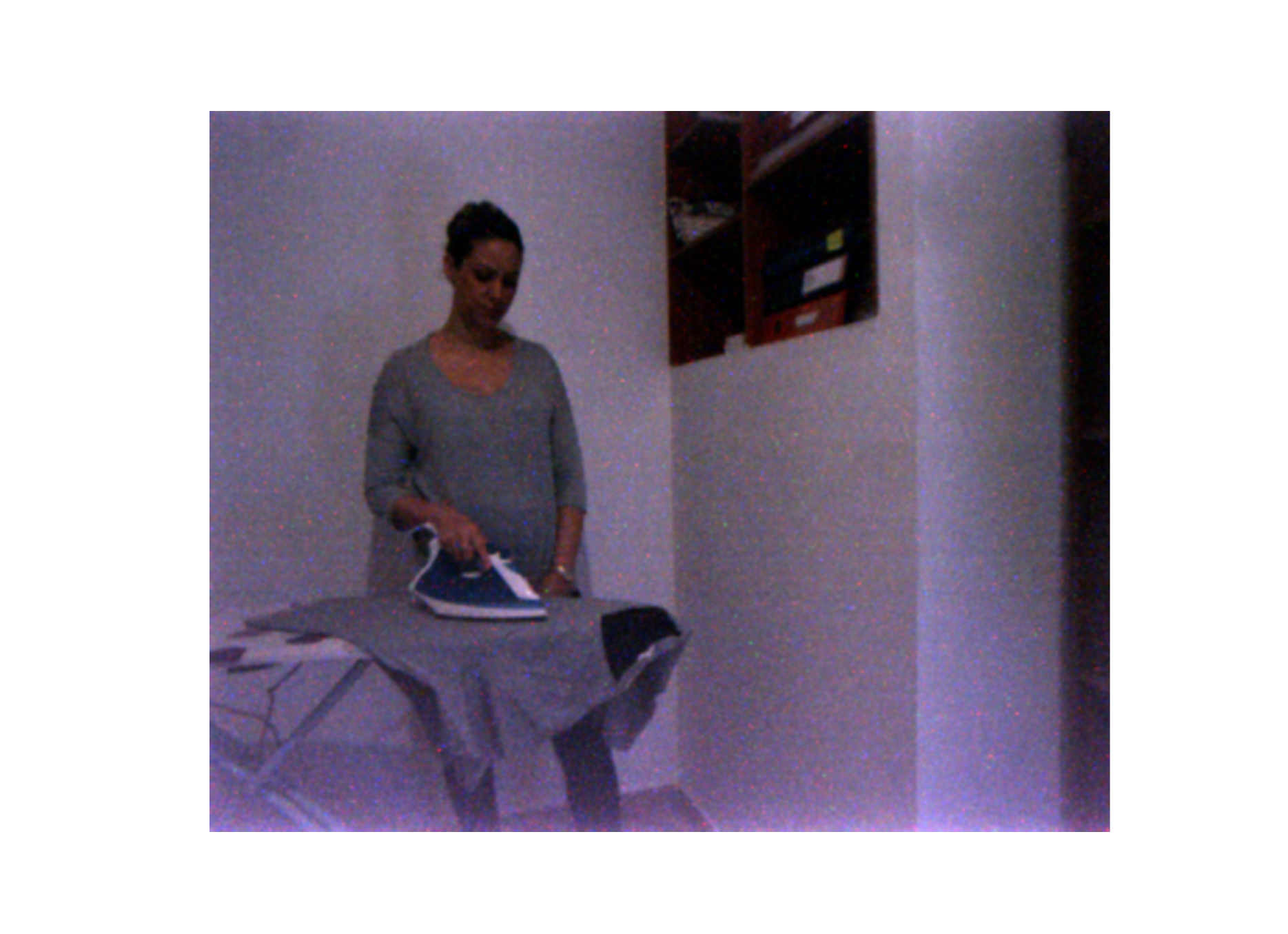}
  \caption{Frame example from \textit{``ironing"} video used for training over-the-air scene based attack.}
  \label{fig:ota_scene_based_train}
\end{subfigure}%
\hspace{5mm}
\begin{subfigure}{0.45\columnwidth}
  \centering
  \includegraphics[width=\columnwidth,trim=0.7cm 0cm 1cm 0.7cm,clip]{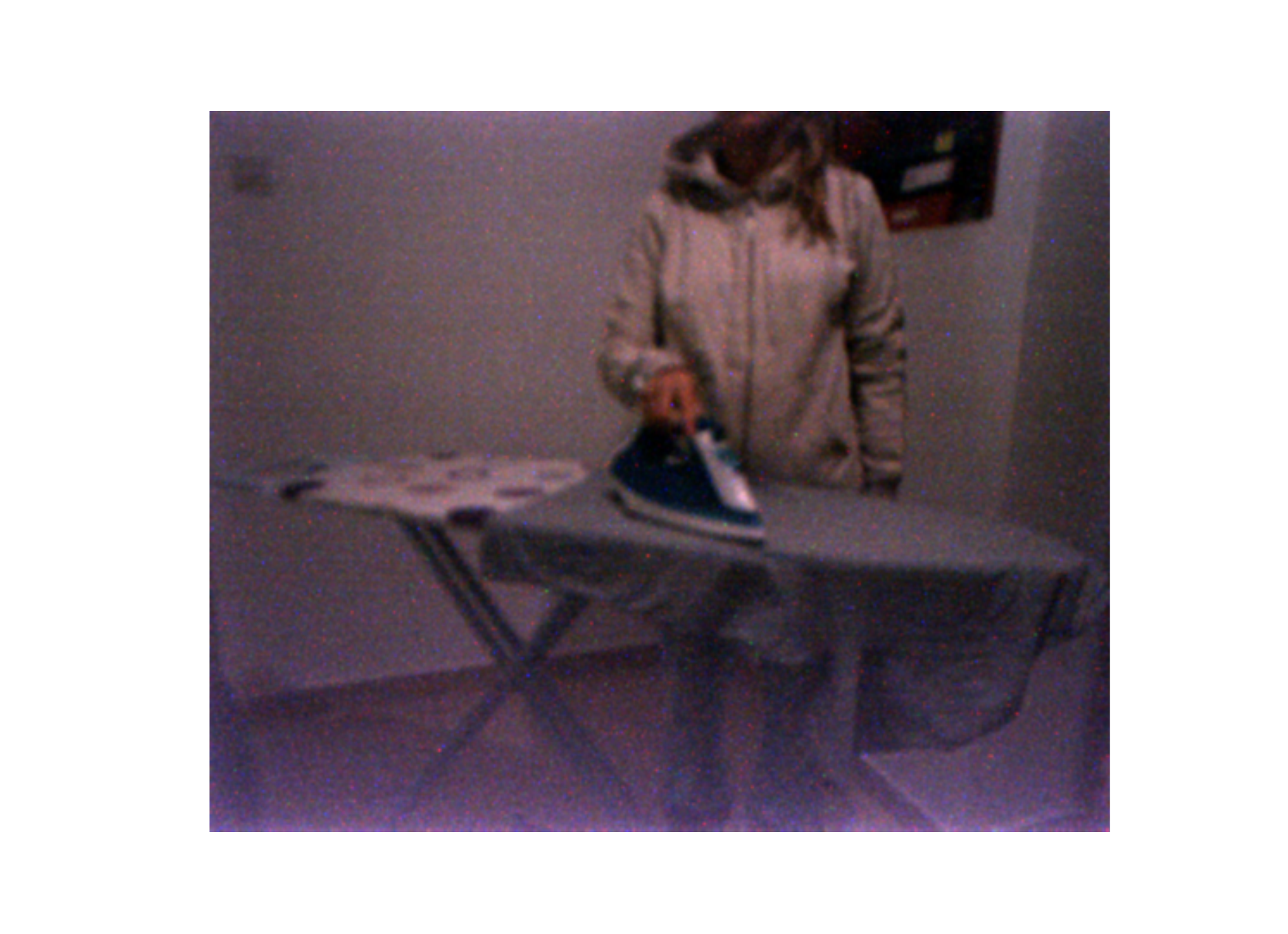}
  \caption{Frame example from \textit{``ironing"} scene used for testing over-the-air scene based attack.}
  \label{fig:ota_scene_based_test}
\end{subfigure}

\caption{Two frames from "similar" scenes.}
\label{fig:ota_scene_based_frames}
\end{adjustbox}
\end{figure*}

\subsection{Over-the-Air Scene-based Flickering Attack}
As described in the paper, in the scene-based approach we assume a prior knowledge of the scene and the action.
In this approach we record a video without any adversarial perturbation of the scene we would like to attack. Then we develop a time-invariant digital attack for this recording as described in the paper.
Once we have the digital attack, we transmit it to a "similar" scene in order to apply the attack in the real world as can be found here\footnote{\url{https://bit.ly/Over_the_Air_scene_based_videos}}.
For illustrating the meaning of "similar" scene, we show in Figure \ref{fig:ota_scene_based_frames} two frames, where Figure \ref{fig:ota_scene_based_train} is a frame example from the video (scene) which the attack was trained upon and Figure \ref{fig:ota_scene_based_test} is a frame example from the scene on which the developed attack was applied on. The relevant videos shows that even though the positioning is different and the clothing are not the same, the attack is still very effective even with a small perturbation.


\section*{Acknowledgements}
Cartoon in Figure \ref{fig:ota_room} designed by "brgfx / Freepik".

\end{appendices}

\end{document}